\pdfoutput=1

\documentclass[11pt]{article}
\usepackage{hyperref}
\usepackage{multirow}
\usepackage{graphicx}

\usepackage[final]{acl}

\usepackage{times}
\usepackage{latexsym}

\usepackage[T1]{fontenc}

\usepackage[utf8]{inputenc}

\usepackage{microtype}

\usepackage{inconsolata}
\usepackage{algorithm}
\usepackage{algorithmic}
\usepackage{amsmath}
\usepackage{amssymb}
\usepackage{amsthm}
\usepackage{caption}
\usepackage{graphicx}
\usepackage{booktabs}
\usepackage{colortbl}
\usepackage{arydshln}
\usepackage{varwidth}

\newtheorem*{definition}{Definition (Token Cohesiveness)}
\newtheorem*{hypothesis}{Hypothesis (Token Cohesiveness Disparity)}

\title{Zero-Shot Detection of LLM-Generated Text using Token Cohesiveness}

\author{Shixuan Ma \and Quan Wang\thanks{Corresponding author: Quan Wang.}\\
MOE Key Laboratory of Trustworthy Distributed Computing and Service, \\
Beijing University of Posts and Telecommunications \\
\texttt{\{1132685456, wangquan\}@bupt.edu.cn}\\
}

\begin{document}
\maketitle
\begin{abstract}
The increasing capability and widespread usage of large language models (LLMs) highlight the desirability of automatic detection of LLM-generated text. Zero-shot detectors, due to their training-free nature, have received considerable attention and notable success. In this paper, we identify a new feature, token cohesiveness, that is useful for zero-shot detection, and we demonstrate that LLM-generated text tends to exhibit higher token cohesiveness than human-written text. Based on this observation, we devise TOCSIN, a generic dual-channel detection paradigm that uses token cohesiveness as a plug-and-play module to improve existing zero-shot detectors. To calculate token cohesiveness, TOCSIN only requires a few rounds of random token deletion and semantic difference measurement, making it particularly suitable for a practical black-box setting where the source model used for generation is not accessible. Extensive experiments with four state-of-the-art base detectors on various datasets, source models, and evaluation settings demonstrate the effectiveness and generality of the proposed approach. Code available at: \url{https://github.com/Shixuan-Ma/TOCSIN}.
\end{abstract}

\section{Introduction}
The past few years have witnessed tremendous advances in Large Language Models (LLMs). These models such as ChatGPT~\cite{openai2023chatgpt}, PaLM \cite{chowdhery2023palm}, GPT-4 \cite{achiam2023gpt} can now generate text of supreme quality, demonstrating exceptional performance in various fields like question answering, news reporting, and story writing. The increasing capability of LLMs to produce human-like text at high efficiency, however, also raises concerns about their misuse for malicious purposes, e.g., phishing \cite{panda2024teach}, disinformation \cite{jiang2024disinformation}, and academic dishonesty \cite{perkins2023academic}. The effective detection of LLM-generated text therefore becomes a vital principle to ensure the responsible use of LLMs.

LLM-generated text detection is typically formulated as a binary classification task, i.e., to classify if a piece of text is generated by a particular source LLM or written by human \cite{tang2024science}. Current solutions roughly fall into two categories: supervised classifiers and zero-shot classifiers. Supervised classifiers are trained from labeled data and thus may overfit to their specific training domains \cite{wang2023m4}. Zero-shot classifiers, in contrast, are entirely training-free, making them less prone to domain-specific degradation and typically generalizing better \cite{zhu2023beat}.

\begin{figure}[t]
  \centering
  \includegraphics[width=0.48\textwidth]{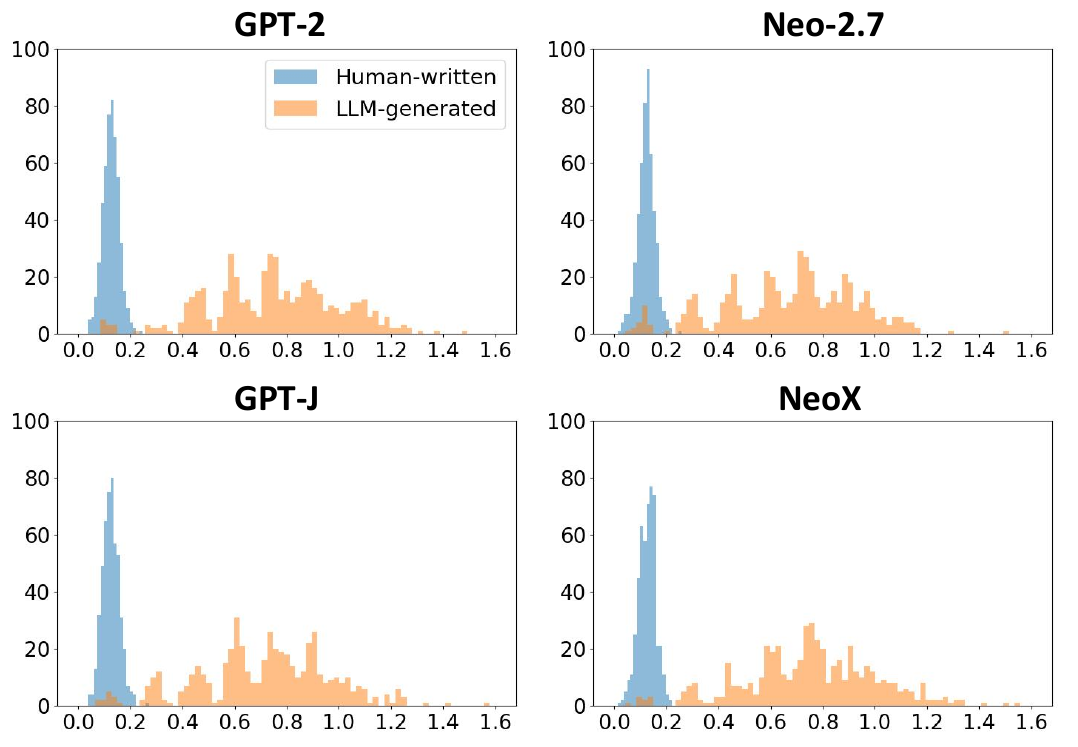}
  \caption{Histograms of token cohesiveness distributions for 500 human-written and 500 LLM-generated articles. Human-written articles are sampled from XSum \cite{narayan-etal-2018-dont}, and LLM-generated articles are produced by prompting four source models with the first 30 tokens of each human-written article. The calculation of token cohesiveness will be detailed in Section~\ref{subsec:assertion}.}
  \label{fig:example1}
\end{figure}

\begin{figure*}[t]
  \centering
  \includegraphics[width=1\textwidth]{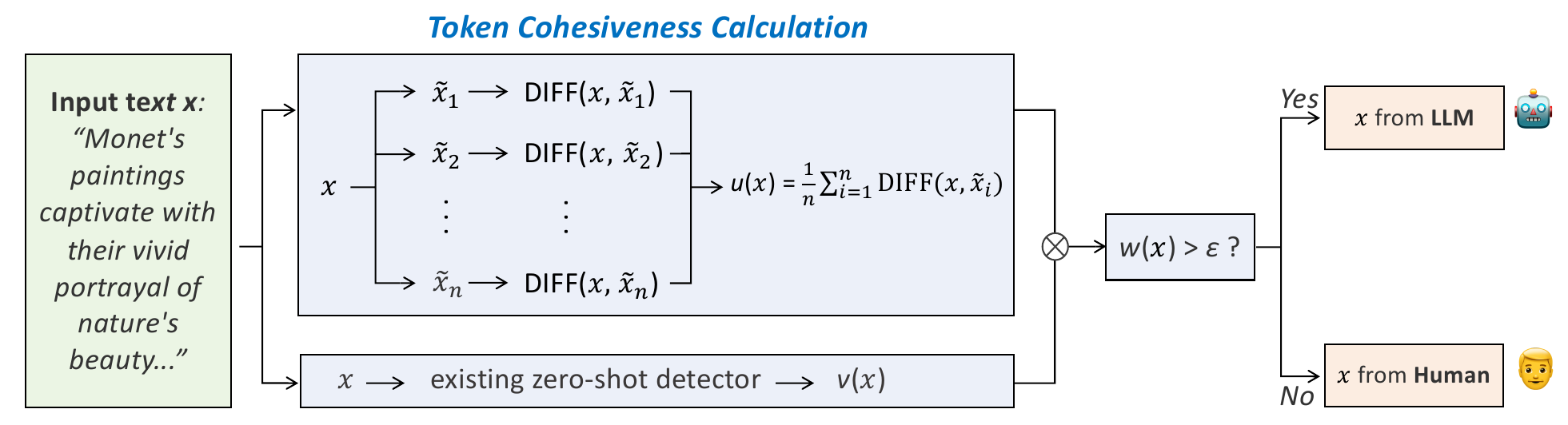}
  \caption{Overview of TOCSIN. The input text $x$ is fed into the upper channel to calculate token cohesiveness $u(x)$, and the lower channel to produce raw prediction $v(x)$. The two scores are then combined into $w(x)$, and if the combination exceeds a predefined threshold $\epsilon$, the text $x$ is categorized as LLM-generated.}
  \label{fig:example2}
\end{figure*}

Most existing zero-shot detectors are developed based on the generation probabilities (or their variations) of the source model, assuming that LLM-generated text aligns better with these probabilities \cite{gehrmann2019gltr,mitchell2023detectgpt}. This paper takes a different tack and introduces a fundamentally new feature, {\bf token cohesiveness}, which does not rely on the source model's output to detect LLM-generated text. Token cohesiveness is defined as the expected semantic difference between input text $x$ and its copy $\tilde{x}$ after randomly removing a small proportion of tokens. It essentially measures how closely the removed tokens are semantically related to the rest of the input text. Then our key assertion is that \textbf{LLM-generated text generally exhibits higher token cohesiveness than human-written text}. This is because LLMs use the causal self-attention mechanism \cite{NIPS2017_3f5ee243} to generate text, requiring the generation of each token to be conditioned on all its preceding tokens, which would naturally foster a closer relationship among tokens, thus increasing token cohesiveness. In contrast, humans tend to write text more freely with no such explicit restriction, which potentially results in a looser relationship among tokens, thus reducing token cohesiveness. We empirically verify this assertion, and find that it holds true across a diverse body of LLMs, as illustrated in Figure~\ref{fig:example1}.

Given the discriminative power of the new feature and its distinctiveness from existing detectors, we propose \textbf{TOCSIN}, a novel paradigm that leverages TOken CoheSIveNess to enhance zero-shot detection of LLM-generated text. TOCSIN is a dual-channel detector, with one channel equipped with an existing zero-shot detector, and the other channel a token cohesiveness calculation module. Given a piece of text to be detected, we create multiple copies, with each copy randomly removing a specific proportion of tokens from the input text. We calculate the average semantic difference, or more precisely, the average negative BARTScore \cite{NEURIPS2021_e4d2b6e6} between the input text and these copies as the token cohesiveness score. Meanwhile, we feed the input text into an existing zero-shot detector to produce a raw prediction score. After that, we combine the two scores and perform thresholding on the combination to make the final decision. The overall procedure is sketched in Figure~\ref{fig:example2}.  

As an enhancement to existing zero-shot detectors, TOCSIN enjoys several merits. (1) \textbf{General applicability:} The dual-channel solution of TOCSIN is quite generic, allowing token cohesiveness to be used as a plug-and-play module in a  variety of detectors. (2) \textbf{Low additional time cost:} TOCSIN keeps the existing detector channel unchanged, and the additional time cost mainly comes from token cohesiveness calculation, which only involves several rounds of random token deletion and BART-Score computation, and is highly time efficient. (3) \textbf{Low additional space cost:} The only additional space cost comes from loading BART \cite{lewis-etal-2020-bart} (or more precisely, BART-base), which is relatively small compared to those scoring models used in existing zero-shot detectors.

To rigorously evaluate the effectiveness and generality of TOCSIN, we apply it to four current state-of-the-art zero-shot detectors, Likelihood \cite{mitchell2023detectgpt}, LogRank \cite{mitchell2023detectgpt}, LRR \cite{su2023detectllm} and Fast-DetectGPT \cite{bao2024fastdetectgpt}, and conduct extensive experiments on four diversified datasets, with LLM-generated passages produced by eight different source models ranging from GPT-2 \cite{radford2019language} to GPT-4 \cite{achiam2023gpt} and Gemini \cite{team2023gemini}. Experimental results demonstrate consistent and meaningful improvements over the four detectors in both white-box and black-box settings.

Our main contributions in this paper are three-fold: (1) unveiling and validating a new hypothesis that LLM-generated text exhibits higher token cohesiveness than human-written text, (2) proposing a novel and generic framework that uses token cohesiveness to improve zero-shot detection of LLM-generated text, and (3) achieving new best detection accuracy compared to existing zero-shot detectors.

\section{Related Work}
The increasingly powerful LLMs \cite{radford2019language,openai2023chatgpt,achiam2023gpt,chowdhery2023palm,touvron2023llama}, though demonstrating excellent performance on various language-related tasks, raise numerous ethical concerns, drawing extensive attention to automatic detection of LLM-generated text \cite{guo2023close,li2023deepfake}. 

LLM-generated text detection is typically formulated as a binary classification task, with current solutions roughly categorized into supervised classifiers and zero-shot classifiers. Supervised classifiers are those trained with statistical features \cite{solaiman2019release,ippolito2020automatic,wu2023llmdet,verma2023ghostbuster} or neural representations \cite{uchendu2020authorship,bakhtin2019real,zhong2020neural,bhattacharjee-etal-2023-conda,wang-etal-2024-ideate-detecting} to discriminate LLM-generated and human-written text, wherein a popular trend is to directly fine-tune a pre-trained language model like BERT \cite{devlin-etal-2019-bert} or RoBERTa \cite{liu2019roberta} for the classification task \cite{NEURIPS2019_3e9f0fc9,rodriguez-etal-2022-cross,mitrovic2023chatgpt,chen2023gpt}. These supervised classifiers, though achieving excellent performance on their training domains, require periodic retraining to adapt to new LLMs, and often exhibit a tendency to overfit their training data \cite{pu2023deepfake}.

Zero-shot classifiers are entirely training-free and often show better generalization ability. Their key ideas are to extract various statistical features, e.g., likelihood \cite{hashimoto-etal-2019-unifying}, perplexity \cite{lavergne2008detecting}, normalized log-rank perturbation \cite{su2023detectllm}, probability and conditional probability curvatures \cite{mitchell2023detectgpt,bao2024fastdetectgpt}, and perform thresholding on these features to discern LLM-generated and human-written text. Such features can be collected from the source LLMs themselves (the white-box setting) or from surrogate models (the black-box setting). In this paper, we propose a new feature called token cohesiveness to improve existing zero-shot detectors. The computation of the new feature does not rely on the source LLMs, making it particularly suitable for the black-box setting.

Recently there emerge some black-box zero-shot detectors based on LLM rewriting \cite{zhu2023beat,yang2024dnagpt}. The idea is to rewrite a passage using another LLM, typically ChatGPT, and then assess the similarity or overlap between the original and recomposed text. Passages with larger similarity or overlap will be regarded as LLM-generated. Our approach similarly adheres to the principle of text reconstruction. But it considers only random token deletion and does not require additional API calls, thus is more efficient and economical.

Besides binary classification, some recent studies consider more challenging LLM-generated text detection tasks, e.g., detecting mixtures of human-written and LLM-modified text \cite{wang-etal-2023-seqxgpt} and tracing the origin of text generation \cite{li2023origin}. These topics are out of the scope of this paper and will be studied as future work. 

\section{Methodology}\label{sec:method}
This section presents TOCSIN, a novel paradigm to improve zero-shot detection of LLM-generated text. Below we formally define the task in Section~\ref{subsec:task}, then illustrate our key assumption in Section~\ref{subsec:assertion}, followed by the detailed approach in Section~\ref{subsec:tocsin}.

\subsection{Problem Formulation}\label{subsec:task}
We study zero-shot LLM-generated text detection, which is formulated as a binary classification problem. Given a piece of text, or candidate passage $x$, the goal is to discern whether $x$ is human-written or generated by a source LLM. The problem is zero-shot in the sense that we do not assume access to any labeled samples to perform detection. 

The method we propose is an enhancement to existing zero-shot detectors. Besides the requirements of the base detector, it also makes use of a semantic similarity measurement model, e.g., BART-Score \cite{NEURIPS2021_e4d2b6e6}, to calculate token cohesiveness of the candidate passage (see Section~\ref{subsec:assertion} for details). This model is relatively small and is used off-the-shelf without any fine-tuning.

\subsection{Key Assumption}\label{subsec:assertion}
We introduce a new feature, token cohesiveness, to distinguish between LLM-generated and human-written text, and our key assumption is that samples from a source LLM typically exhibit higher token cohesiveness than human-written text. Below we formally define token cohesiveness and state the assumption with an empirical verification.

\begin{definition}
Given a candidate passage $x$, let $\tilde{x}$ denote a random copy created by removing a small proportion of tokens from $x$. The token cohesiveness of $x$ is then defined as the expectation of semantic difference between $x$ and $\tilde{x}$, i.e., $u(x)\triangleq\mathbb{E}(\mathrm{DIFF}(x,\tilde{x}))$, where $\mathrm{DIFF}(\cdot,\cdot)$ is a semantic difference metric, and $\mathbb{E}(\cdot)$ the expectation operation.
\end{definition}

Token cohesiveness essentially measures the semantic closeness among tokens in the passage. The closer the tokens are semantically related to each other, the higher the token cohesiveness would be. We argue that there is a gap in token cohesiveness between LLM-generated and human-written text. For LLM-generated text, each token is generated based on all its preceding tokens. This would naturally foster a closer relationship among tokens and lead to higher token cohesiveness. But for human-written text, there is no explicit restriction about token generation, potentially resulting in a looser relationship among tokens and, consequently, lower token cohesiveness. We formalize the assertion as a token cohesiveness disparity hypothesis.

\begin{hypothesis}
Let $\mathcal{P}_{LLM}$ denote the distribution of LLM-generated text, and $\mathcal{P}_{Human}$ that of human-written text. Then the token cohesiveness $u(x)$ tends to be higher for samples $x \sim \mathcal{P}_{LLM}$, while lower for $x \sim \mathcal{P}_{Human}$.
\end{hypothesis}

We empirically verify the hypothesis in an automated manner. Specifically, as in prior work \cite{bao2024fastdetectgpt,mitchell2023detectgpt}, we use 500 news articles randomly sampled from XSum \cite{narayan-etal-2018-dont} as human-written data, and use the output of four different LLMs when prompted with the first 30 tokens of each human-written article as LLM-generated data. For each article, to calculate its token cohesiveness, we create 10 copies, each randomly deleting 1.5\% tokens from the original article. We use negative BARTScore \cite{NEURIPS2021_e4d2b6e6} as the semantic difference metric $\mathrm{DIFF}(\cdot,\cdot)$, and approximate the expectation $\mathbb{E}(\cdot)$ with the average of the 10 copies. Figure~\ref{fig:example1} shows the results, revealing that the token cohesiveness distributions do differ significantly between LLM-generated and human-written data. LLM-generated samples typically show a broader distribution with higher token cohesiveness values.

\subsection{Detailed Approach}\label{subsec:tocsin}
Based on the above findings, we devise TOCSIN, a generic dual-channel detection paradigm that uses token cohesiveness as a plug-and-play module to improve existing zero-shot detectors. The overall architecture of TOCSIN is sketched in Figure~\ref{fig:example2}. 

Specifically, given a passage $x$, we first feed it into one channel to calculate its token cohesiveness. This channel creates $n$ copies $\{\tilde{x}_1, \tilde{x}_2, \cdots, \tilde{x}_n\}$ for the input $x$, with each copy randomly deleting a certain proportion (denoted as $\rho$) of tokens from $x$. It then estimates the token cohesiveness of $x$ as:
\begin{equation}
u(x) = \sum_{i=1}^{n}\frac{\mathrm{DIFF}(x,\tilde{x}_i)}{n},
\end{equation}
where $\mathrm{DIFF}(\cdot,\cdot)$ measures the semantic difference between $x$ and each of its copies $\tilde{x}_i$. We employ the established negative BARTScore \cite{NEURIPS2021_e4d2b6e6} as the metric, with other evaluated metrics discussed in Appendix~\ref{appendix:similarity-metrics}. Meanwhile, we feed the input $x$ into another channel, which is equipped with an existing zero-shot detector, to produce a raw prediction $v(x)$. The output of the two channels are then combined into a new prediction:
\begin{equation}
w(x) = \left\{ \begin{array}{lcl}
e^{u(x)} \times v(x) & \mbox{if} & v(x) \geq 0, \\
e^{-u(x)} \times v(x) & \mbox{if} & v(x) < 0,
\end{array}\right.
\end{equation}
on which we perform thresholding to make the final decision, i.e., $x$ is categorized as LLM-generated if $w(x) > \epsilon$ or otherwise human-written. This detection procedure is summarized into Algorithm~\ref{alg:example1}.
       
\begin{algorithm}[t]\small
\caption{\small TOCSIN LLM-generated text detection}
\label{alg:example1}
\begin{algorithmic}[1]
    \REQUIRE passage $x$, base detector $\mathrm{BASE}(\cdot)$, random token deletion operation $\mathrm{RTD}(\cdot)$, semantic difference metric $\mathrm{DIFF}(\cdot, \cdot)$, decision threshold $\epsilon$
    \ENSURE True – LLM-generated, False – human-written
    \STATE $\{\tilde{x}_i\}_{i=1}^n \gets \mathrm{RTD}(x)$ for $i=1,\cdots,n$ \textcolor{gray}{\COMMENT{create copies}}
    \STATE $u(x) \gets \sum_{i=1}^{n}\frac{\mathrm{DIFF}(x,\tilde{x}_i)}{n}$ \textcolor{gray}{\COMMENT{token cohesiveness}}
    \STATE $v(x) \gets \mathrm{BASE}(x)$ \textcolor{gray}{\COMMENT{raw prediction}}
    \STATE $w(x) \gets \left\{ \begin{array}{lcl}
\negthickspace\negmedspace e^{u(x)} \times v(x) & \negthickspace\negthickspace\mbox{if}\negthickspace\negthickspace\negthickspace & v(x) \geq 0 \\
\negthickspace\negmedspace e^{-u(x)} \times v(x) & \negthickspace\negthickspace\mbox{if}\negthickspace\negthickspace\negthickspace & v(x) < 0
\end{array}\right.$ \textcolor{gray}{\COMMENT{combination}}
    \STATE \textbf{return} $w(x) > \epsilon$
\end{algorithmic}
\end{algorithm}

The proposed dual-channel detection paradigm is rather generic, allowing token cohesiveness to be applied as a plug-and-play module in a variety of detectors to further improve their performance. This paper chooses four state-of-the-art base detectors, including Likelihood \cite{mitchell2023detectgpt}, LogRank \cite{mitchell2023detectgpt}, LRR \cite{su2023detectllm} and Fast-DetectGPT \cite{bao2024fastdetectgpt}, the details of which are provided in Appendix~\ref{appendix:base-detector}.

\section{Experiments}\label{sec:EXP}
This section evaluates the effectiveness of TOCSIN for zero-shot LLM-generated text detection compared with prior state-of-the-arts, and also provides extensive additional analyses to better understand multiple facets of the proposed method.

\subsection{Experimental Setups}
\paragraph{Datasets}
To ensure fair comparison, we follow prior work~\cite{bao2024fastdetectgpt} to use four diversified datasets: \emph{XSum} for news articles \cite{narayan-etal-2018-dont}, \emph{SQuAD} for Wikipedia content \cite{rajpurkar-etal-2016-squad}, \emph{WritingPrompts} for storytelling \cite{fan-etal-2018-hierarchical}, and \emph{PubMedQA} for biomedical question answering \cite{jin-etal-2019-pubmedqa}. Each dataset contains 150 to 500 randomly sampled human-written passages as negative samples, and the same number of LLM-generated passages as positive samples, created by prompting a source model with the first 30 tokens of each negative sample. Eight different source models of various size are considered, including the 1.5B \emph{GPT-2} \cite{radford2019language}, 2.7B \emph{OPT-2.7} \cite{zhang2022opt}, 2.7B \emph{GPT-Neo-2.7} \cite{gpt-neo}, 6B \emph{GPT-J} \cite{gpt-j}, 20B \emph{GPT-NeoX} \cite{black-etal-2022-gpt}, as well as OpenAI's most powerful \emph{ChatGPT} \cite{openai2023chatgpt}, \emph{GPT-4} \cite{achiam2023gpt} and Google's most powerful \emph{Gemini} \cite{team2023gemini} to simulate text generation in real-world scenarios. All these datasets are collected from the open-source project of Fast-DetectGPT \cite{bao2024fastdetectgpt}, except for those generated by Gemini, the details of which are introduced in Appendix~\ref{appendix:Gemini}.

\begin{table*}[t]
  \centering\small
  \begin{tabular*}{1 \textwidth}{@{\extracolsep{\fill}}lcccccc}
    \toprule
    {\bf Method} & {\bf GPT-2} & {\bf OPT-2.7} & {\bf Neo-2.7} & {\bf GPT-J} & {\bf NeoX} & {\bf Avg.} \\
    \midrule
    \rowcolor{gray!50}
    \multicolumn{7}{c}{\textbf{The White-Box Setting}}\\
    Entropy \cite{mitchell2023detectgpt}         & 0.5174 & 0.4830 & 0.4898 & 0.5005 & 0.5333 & 0.5048 \\
    DNA-GPT \cite{yang2024dnagpt}$\dag$ & 0.9024 & 0.8797 & 0.8690 & 0.8227 & 0.7826 & 0.8513 \\ 
    DetectGPT \cite{mitchell2023detectgpt}$\dag$ & 0.9917 & 0.9758 & 0.9797 & 0.9353 & 0.8943 & 0.9554 \\    
    NPR \cite{su2023detectllm}$\dag$                    & 0.9948 & 0.9832 & 0.9883 & 0.9500 & 0.9065 & 0.9645 \\
    \hdashline
    Likelihood \cite{mitchell2023detectgpt}     & 0.9125 & 0.8963 & 0.8900 & 0.8480 & 0.7946 & 0.8683 \\  
    Likelihood+TOCSIN (ours)     & {\bf 0.9905}&{\bf 0.9876}  & {\bf 0.9794} &{\bf  0.9776 }& {\bf 0.9549} &{\bf  0.9780 }\\
    {\it(Absolute $\uparrow$)} & {\it 7.80\%}  & {\it 9.13\%}  & {\it 8.94\%}  & {\it 12.96\%} & {\it 16.03\%} & {\it 10.97\%} \\
    \hdashline
    LogRank \cite{mitchell2023detectgpt}       & 0.9385 & 0.9223 & 0.9226 & 0.8818 & 0.8313 & 0.8993 \\ 
    LogRank+TOCSIN (ours)       & {\bf 0.9933} &{\bf 0.9907}  & {\bf 0.9857} & {\bf 0.9811} & {\bf 0.9586} &{\bf 0.9819}
  \\ 
    {\it(Absolute $\uparrow$)} & {\it 5.48\%}  & {\it 6.84\%}  & {\it 6.31\%}  & {\it 9.93\%} & {\it 12.73\%} & {\it8.26\%} \\
    \hdashline
    LRR \cite{su2023detectllm}  & 0.9601 & 0.9401 & 0.9522 & 0.9179 & 0.8793 & 0.9299 \\ 
    LRR+TOCSIN (ours)       & {\bf 0.9919} & {\bf 0.9929} & {\bf 0.9873} & {\bf 0.9914} & {\bf 0.9799} & {\bf 0.9887} \\
    {\it(Absolute $\uparrow$)} & {\it 3.18\%}  & {\it 5.28\%}  & {\it 3.51\%}  & {\it 7.35\%} & {\it 10.06\%} & {\it 5.88\%} \\
    \hdashline
    Fast-DetectGPT \cite{bao2024fastdetectgpt}  & 0.9967 & 0.9908 & 0.9940 & 0.9866 & 0.9754 & 0.9887 \\ 
    Fast-DetectGPT+TOCSIN (ours) & {\bf 0.9986} & {\bf 0.9960} & {\bf 0.9978} & {\bf 0.9941} & {\bf 0.9863} & {\bf 0.9946} \\
    {\it(Absolute $\uparrow$)} & {\it 0.19\%} & {\it 0.52\%} & {\it 0.38\%} & {\it 0.75\%} & {\it 1.09\%} & {\it 0.59\%} \\
    \rowcolor{gray!50}
    \multicolumn{7}{c}{\textbf{The Black-Box Setting}}\\
    DetectGPT \cite{mitchell2023detectgpt}$\dag$ & 0.8517 & 0.8390 & 0.9797 & 0.8575 & 0.8400 & 0.8736 \\
    \hdashline
    Likelihood (ours)     & 0.7625 &0.7838  &0.8899  &0.8054  & 0.7851 &0.8053  \\  
    Likelihood+TOCSIN (ours)     & {\bf 0.9626}&{\bf 0.9723 } &{\bf 0.9794}  & {\bf 0.9722} & {\bf 0.9628} &{\bf  0.9699} \\
    {\it(Absolute $\uparrow$)} & {\it 20.01\%}  & {\it 18.85\%}  & {\it 8.95\%}  & {\it 16.68\%} & {\it 17.77\%} & {\it 16.46\%} \\
    \hdashline
     LogRank (ours)       &  0.8013&0.8210  & 0.9226 & 0.8362 & 0.8070 & 0.8376 \\ 
    LogRank+TOCSIN (ours)       & {\bf 0.9644 }&{\bf 0.9753}  &{\bf 0.9857 } &{\bf 0.9737}  &{\bf 0.9630} & {\bf 0.9724} \\ 
    {\it(Absolute $\uparrow$)} & {\it 16.31\%}  & {\it 15.43\%}  & {\it 6.31\%}  & {\it 13.75\%} & {\it 15.60\%} & {\it 13.48\%} \\
    \hdashline
    LRR (ours) & 0.8505 & 0.8609 & 0.9518 & 0.8637 & 0.8187 & 0.8691 \\
    LRR+TOCSIN (ours) & {\bf 0.9745} & {\bf 0.9849} & {\bf 0.9873} & {\bf 0.9832} & {\bf 0.9752} & {\bf 0.9810} \\
    {\it(Absolute $\uparrow$)} & {\it 12.40\%} & {\it 12.40\%} & {\it 3.55\%} & {\it 11.95\%} & {\it 15.65\%} & {\it 11.19\%} \\
    \hdashline
    Fast-DetectGPT \cite{bao2024fastdetectgpt} & 0.9834 & 0.9572 & 0.9984 & 0.9592 & 0.9404 & 0.9677 \\
    Fast-DetectGPT+TOCSIN (ours) & {\bf 0.9948} & {\bf 0.9815} & {\bf 0.9994} & {\bf 0.9822} & {\bf 0.9741} & {\bf 0.9864} \\
    {\it(Absolute $\uparrow$)} & {\it 1.14\%} & {\it 2.43\%} & {\it 0.10\%} & {\it 2.30\%} & {\it 3.37\%} & {\it 1.87\%} \\
    \bottomrule
  \end{tabular*}
  \caption{
AUROC for zero-shot detection of passages generated from five source models, averaged across XSum, SQuAD, WritingPrompts, with detailed results provided in Appendix~\ref{subsec:open-source-models}. Results marked by ``(ours)'' are produced by ourselves, and other results are taken directly from \cite{bao2024fastdetectgpt} to avoid implementation bias. In the white-box (resp. black-box) setting, the source model (resp. GPT-Neo-2.7) is used for scoring. $\dag$ denotes methods that invoke scoring models multiple times, thereby significantly increasing computational demands. {\bf Bold} indicates that a +TOCSIN variant outperforms its direct baseline, and ``{\it(Absolute $\uparrow$)}'' the absolute improvements.
 }
  \label{tab:open-source-models}
\end{table*}

\begin{table*}[t]
  \centering\small
  \setlength{\tabcolsep}{3pt}
  \begin{tabular*}{1.0 \textwidth}{@{\extracolsep{\fill}}lccccccccc}
    \toprule
    \multirow{2}{*}{\bf Method} & \multicolumn{3}{c}{\textbf{ChatGPT}} & \multicolumn{3}{c}{\textbf{GPT-4}} & \multicolumn{3}{c}{\textbf{Gemini}}\\
    \cmidrule(r){2-4}\cmidrule(l){5-7}\cmidrule(l){8-10}
    & {\bf XSum} & {\bf Writing} & {\bf PubMed}  & {\bf XSum} & {\bf Writing} & {\bf PubMed} & {\bf XSum} & {\bf Writing} & {\bf PubMed}  \\
    \midrule
    \rowcolor{gray!50}
    \multicolumn{10}{c}{\textbf{Supervised Classifiers}}\\
    RoBERTa-base \cite{bao2024fastdetectgpt} &0.9150 &0.7084 &0.6188  &0.6778 &0.5068 &0.5309 &0.8708& 0.8002&0.4460\\
    RoBERTa-large \cite{bao2024fastdetectgpt} &0.8507 &0.5480 &0.6731  &0.6879 &0.3821 &0.6067 &0.8101&0.6296&0.4508 \\
    GPTzero \cite{tian2023gptzero}                     &0.9952 &0.9292 &0.8799  &0.9815 &0.8262 &0.8482& 0.9987&0.9837 &0.8840  \\
    \rowcolor{gray!50}
    \multicolumn{10}{c}{\textbf{Zero-Shot Classifiers}}\\
    Entropy \cite{mitchell2023detectgpt}     &0.3305 &0.1902 &0.2767  &0.4360 &0.3702 &0.3295  & 0.5399&0.4395 &0.4335\\  
    DNA-GPT \cite{yang2024dnagpt}$\dag$ &0.9124 &0.9425 &0.7959  &0.7347 &0.8032 &0.7565  & 0.8675& 0.9257&0.5199 \\
    DetectGPT (ours)$\dag$ &0.8901 &0.9452 &0.6362  &0.6692 &0.8177 &0.5927   &0.7549 &0.9151 &0.6854\\ 
    NPR \cite{su2023detectllm}$\dag$ &0.7899 &0.8924 &0.6784  &0.5280 &0.6122 &0.6328   & 0.8172&0.9487 &0.6384\\
    \hdashline
    Likelihood \cite{mitchell2023detectgpt} & 0.9578 &  0.9740& {\bf 0.8775}  & 0.7980 & 0.8553 &  {\bf 0.8104 } & 0.8519&0.9114 &0.7616\\ 
    Likelihood+TOCSIN (ours) &{\bf 0.9984} &{\bf 0.9933} &  0.8701 & {\bf 0.9736}&{\bf 0.9324} &0.8044  & {\bf 0.8691}& {\bf 0.9256}& {\bf 0.9823}\\ 
    {\it(Absolute $\uparrow$)} &{\it 4.06\%} &{\it 1.93\%} &{\it -0.74\%}  &{\it 17.56\%} &{\it 7.71\%} &{\it -0.60\%}   &{\it 1.72\%} & {\it 1.42\%}&{\it 22.07\%}\\
    \hdashline
    LogRank \cite{mitchell2023detectgpt} &0.9582 &  0.9656&  {\bf 0.8687}& 0.7975& 0.8286&{\bf 0.8003}   & 0.8628&0.9076 &0.7689\\ 
    LogRank+TOCSIN (ours) & {\bf 0.9981}&{\bf 0.9933} &0.8620 &{\bf 0.9716} &{\bf 0.9208} &0.7952  &{\bf 0.8655} &{\bf 0.9175 }& {\bf 0.9716}\\ 
    {\it(Absolute $\uparrow$)} &{\it 3.99\%} &{\it 2.77\%} &{\it -0.67\%}  &{\it 17.41\%} &{\it 9.22\%} &{\it -0.51\%}   &{\it 0.27\%} &{\it 0.99\%} &{\it 20.27\%}\\
    \hdashline
    LRR \cite{su2023detectllm} &0.9162 &0.8958 &{\bf 0.7433}  &0.7447 &0.7028 &{\bf 0.6814}   &0.7274 &0.8179 &0.7234\\ 
    LRR+TOCSIN (ours) &{\bf 0.9939} &{\bf 0.9927} &0.7092  &{\bf 0.9614} &{\bf 0.8036} &0.6465 &{\bf 0.8720} & {\bf 0.9195}&{\bf 0.9978}\\
    {\it(Absolute $\uparrow$)} &{\it 7.77\%} &{\it 9.69\%} &{\it -3.41\%}  &{\it 21.67\%} &{\it 10.08\%} &{\it -3.49\%}  &{\it 14.46\%} &{\it 10.16\%} &{\it 27.44\%} \\
    \hdashline
    Fast-DetectGPT \cite{bao2024fastdetectgpt} &0.9907 &0.9916 &{\bf 0.9021}  &0.9067 &0.9612 &{\bf 0.8503}   &0.8518 &0.9465 &0.8769\\ 
  Fast-DetectGPT+TOCSIN (ours) &{\bf 0.9969} &{\bf 0.9964} &0.9011  &{\bf 0.9455} &{\bf 0.9708} &0.8490  & {\bf 0.8697}&{\bf 0.9484 }&{\bf 0.9799}\\
  {\it(Absolute $\uparrow$)} &{\it 0.62\%} &{\it 0.48\%} &{\it -0.10\%}  &{\it 3.88\%} &{\it 0.96\%} &{\it -0.13\%}   &{\it 1.79\%} &{\it 0.19\%} &{\it 10.30\%} \\
    \bottomrule
  \end{tabular*}
  \caption{
AUROC for detecting passages generated by ChatGPT, GPT-4, and Gemini. For ChatGPT and GPT-4, results marked by ``(ours)'' are produced by ourselves, and other results are taken directly from \cite{bao2024fastdetectgpt} to avoid implementation bias. For Gemini, all results are produced by ourselves. The black-box setting is used for all zero-shot classifiers, with GPT-Neo-2.7 as surrogate model. {\bf Bold} stands for better performance between a baseline and its +TOCSIN version, and ``{\it(Absolute $\uparrow$)}'' the absolute improvements.
  }
  \label{tab:api-based-models}
\end{table*}

\paragraph{Metric}
We also follow prior work to use the area under the receiver operating characteristic curve (\emph{AUROC}) as the evaluation metric. This metric is based on dynamic positive-negative thresholds and does not require specifying a fixed threshold, which is particularly challenging in zero-shot scenarios.

\paragraph{Baselines}
As we have discussed in Section~\ref{subsec:tocsin}, TOCSIN is a generic paradigm that can be applied to various zero-shot detectors to further improve their performance. We choose four zero-shot detectors, \emph{Likelihood} (average log-probability), \emph{LogRank} (average log-rank) \cite{mitchell2023detectgpt}, \emph{LRR} (log-probability log-rank ratio) \cite{su2023detectllm} and \emph{Fast-DetectGPT} (conditional probability curvature) \cite{bao2024fastdetectgpt}, and apply TOCSIN on the four detectors to validate its effectiveness and generality. We choose the four detectors as they are recently proposed, computationally efficient, and report current state-of-the-art performance. The derived methods are denoted as \emph{$\ast$+TOCSIN}. 

Besides the four direct baselines, we also compare with other established zero-shot detectors, including \emph{Entropy} (entropy of predictive distribution) \cite{mitchell2023detectgpt}, \emph{NPR} (normalized log-rank of perturbations) \cite{su2023detectllm}, \emph{Detect-GPT} (probability curvature) \cite{mitchell2023detectgpt}, \emph{DNA-GPT} (divergent n-grams in rewrites) \cite{yang2024dnagpt}. We also perform comparisons to supervised classifiers, including the GPT-2 detectors based on \emph{RoBERTa-base/large} \cite{liu2019roberta} crafted by OpenAI and \emph{GPTZero} \cite{tian2023gptzero}.

\paragraph{Implementation}
TOCSIN is conceptually simple and easy to implement. For the base detectors, we use the code of Likelihood\footnote{https://github.com/eric-mitchell/detect-gpt}, LogRank\footnote{https://github.com/baoguangsheng/fast-detect-gpt/}, LRR\footnote{https://github.com/mbzuai-nlp/DetectLLM} and Fast-DetectGPT\footnote{https://github.com/baoguangsheng/fast-detect-gpt/}, and keep their settings unchanged. For token cohesiveness calculation, there are two hyperparameters: $n$ (the number of copies created for each input) and $\rho$ (the proportion of tokens to be deleted in each copy). We empirically set $n=10$, $\rho=1.5\%$ for \textbf{all experiments without re-tuning}. The impact of the two hyperparameters is discussed in Section~\ref{subsec:analyses}.

\subsection{Experiments with Open-Source LLMs}
We first evaluate TOCSIN in detecting text generated by the five open-source LLMs, from GPT-2 (1.5B) to GPT-NeoX (20B). We use 500 human samples along with 500 LLM samples generated by each of the five models across XSum, SQuAD, and WritingPrompts. By following previous work \cite{bao2024fastdetectgpt}, we consider two evaluation settings: (1) the \emph{white-box setting} where each source model is used to score passages, and (2) the \emph{black-box setting} where the source models are not accessible and a surrogate model, i.e.,  GPT-Neo-2.7, is used to score passages. Note that this scoring is required by the base detectors, not by token cohesiveness calculation. Table~\ref{tab:open-source-models} presents average AUROC across the three datasets in the two settings, with detailed per dataset results reported in Appendix~\ref{subsec:open-source-models}.

\paragraph{Results in White-Box Setting}
As we can see, TOCSIN performs particularly well in this setting. It outperforms the four direct baselines (and all the other baselines), irrespective of which dataset or source model is used. The absolute improvement in average AUROC reaches 10.97\% over Likelihood, 8.26\% over LogRank, and 5.88\% over LRR. For Fast-DetectGPT, which already gets a super high average AUROC of 0.9887, TOCSIN still brings consistent and meaningful improvements, pushing the average AUROC further to 0.9946.

\paragraph{Results in Black-Box Setting}
In this setting we observe similar phenomena. TOCSIN, again, consistently outperforms all the baselines. Notably, the improvements are even more significant than those in the white-box setting (the absolute boosts in average AUROC reach 16.46\%, 13.48\%, 11.19\%, and 1.87\% over the four direct baselines). We speculate this is because the base detectors require the source model to make predictions. In the black-box setting where the source model is not available, they have to use a surrogate model for approximation, which inevitably results in performance degradation. TOCSIN, in contrast, requires no such approximation and can therefore resist the degradation.

\subsection{Experiments with API-based LLMs}
We further evaluate TOCSIN in detecting passages generated by ChatGPT, GPT-4, and Gemini to simulate real-world scenarios. We use 150 positive and 150 negative samples on XSum, WritingPrompts, and PubMedQA. As we are not able to access the source models, we consider the black-box setting, with GPT-Neo-2.7 as the surrogate model. Table~\ref{tab:api-based-models} reports the results, showing that TOCSIN can bring consistent improvements to the four direct baselines in almost all cases. The only exception is the PubMedQA data produced by ChatGPT or GPT-4. In these two cases we find that PubMedQA passages, which consist of only 64 tokens on average, are substantially shorter than passages on the other datasets (e.g., 221 tokens on XSum and 218 tokens on WritingPrompts). For LLM-generated text, the generation of each token is an aggregation process, making the generated token more closely related to its preceding tokens. Passages of shorter length undergo fewer such aggregation processes, which may suppress the closeness of tokens therein and reduce token cohesiveness, making it more difficult to distinguish these passages from human-written ones via token cohesiveness. We will show later in Section~\ref{subsec:analyses} the detailed impact of passage length.

\subsection{Additional Analyses}\label{subsec:analyses}
We provide additional analyses to better understand multiple facets of TOCSIN. We consider only the more practical black-box setting where the source LLMs are not accessible.

\paragraph{Time \& Space Efficiency}
As an improvement to existing zero-shot detectors, TOCSIN keeps the base detector unchanged, and the additional time  and space costs mainly come from the computation of token cohesiveness, which involves 10 rounds of random token deletion and BARTScore computation. To rigorously assess the additional costs, we conduct time and space analysis on XSum, SQuAD, and WritingPrompts where the LLM-generated passages are produced by the five open-source models, and we make comparison between Fast-DetectGPT and Fast-DetectGPT+TOCSIN. All experiments here are conducted on a single NVIDIA A40 GPU with 48GB memory. Table~\ref{tab:time-space} reports the average runtime per instance and GPU memory usage of the two variants, with detailed results further provided in Appendix~\ref{subsec:time-space}. Compared to the base detector, TOCSIN brings a relatively low additional runtime of 0.16s per instance and a relatively low additional GPU memory usage of 4.71GB on average.

\begin{table}[t]
  \centering\small
  \begin{tabular*}{0.48 \textwidth}{@{\extracolsep{\fill}}lcc}
    \toprule
    & {\bf Runtime (s)} & {\bf GPU Memory (GB)} \\
    \midrule
    w/o TOCSIN & 0.31& 23.96 \\
    w/ TOCSIN & 0.47 & 28.67 \\
    {\it(Absolute $\uparrow$)} &{\it 0.16} & {\it ~~4.71} \\
    \bottomrule
  \end{tabular*}
  \caption{
   Runtime per instance and GPU memory usage of w/ and w/o TOCSIN variants of Fast-DetectGPT in black-box setting, averaged across five source models on XSum, SQuAD, and WritingPrompts. ``{\it(Absolute $\uparrow$)}'' means additional time/space cost brought by TOCSIN.
  }
  \label{tab:time-space}
\end{table}

\begin{table*}[t]
  \centering\small
  \begin{tabular*}{1 \textwidth}{@{\extracolsep{\fill}}lcccccc}
    \toprule
    {\bf Method} & {\bf GPT-2} & {\bf OPT-2.7} & {\bf Neo-2.7} & {\bf GPT-J} & {\bf NeoX} & {\bf Avg.} \\
    \midrule
    Fast-DetectGPT & 0.9834 & 0.9572 & 0.9984 & 0.9592 & 0.9404 & 0.9677 \\ 
    Fast-DetectGPT+Likelihood & 0.9696 &  0.9472 & 0.9967 & 0.9522 &  0.9316 &  0.9595 \\
    Fast-DetectGPT+LogRank & 0.9786 &  0.9558 &  0.9982 &0.9589 &  0.9389 &  0.9661 \\
    Fast-DetectGPT+LRR & 0.9833 &  0.9598 &  0.9985 &  0.9612 &  0.9420 &  0.9690 \\
    Fast-DetectGPT+TOCSIN & {\bf 0.9948} & {\bf 0.9815} & {\bf 0.9994} & {\bf 0.9822} & {\bf 0.9741} & {\bf 0.9864} \\
    \bottomrule
  \end{tabular*}
  \caption{AUROC for zero-shot detection of passages generated from five source models in the black-box setting, using GPT-Neo-2.7 as the surrogate model. The results are achieved by combining Fast-DetectGPT with different detectors, and averaged across XSum, SQuAD, and WritingPrompts.
 }
  \label{tab:Ensemble}
\end{table*}

\begin{figure}[t]
  \centering
  \includegraphics[width=0.48\textwidth]{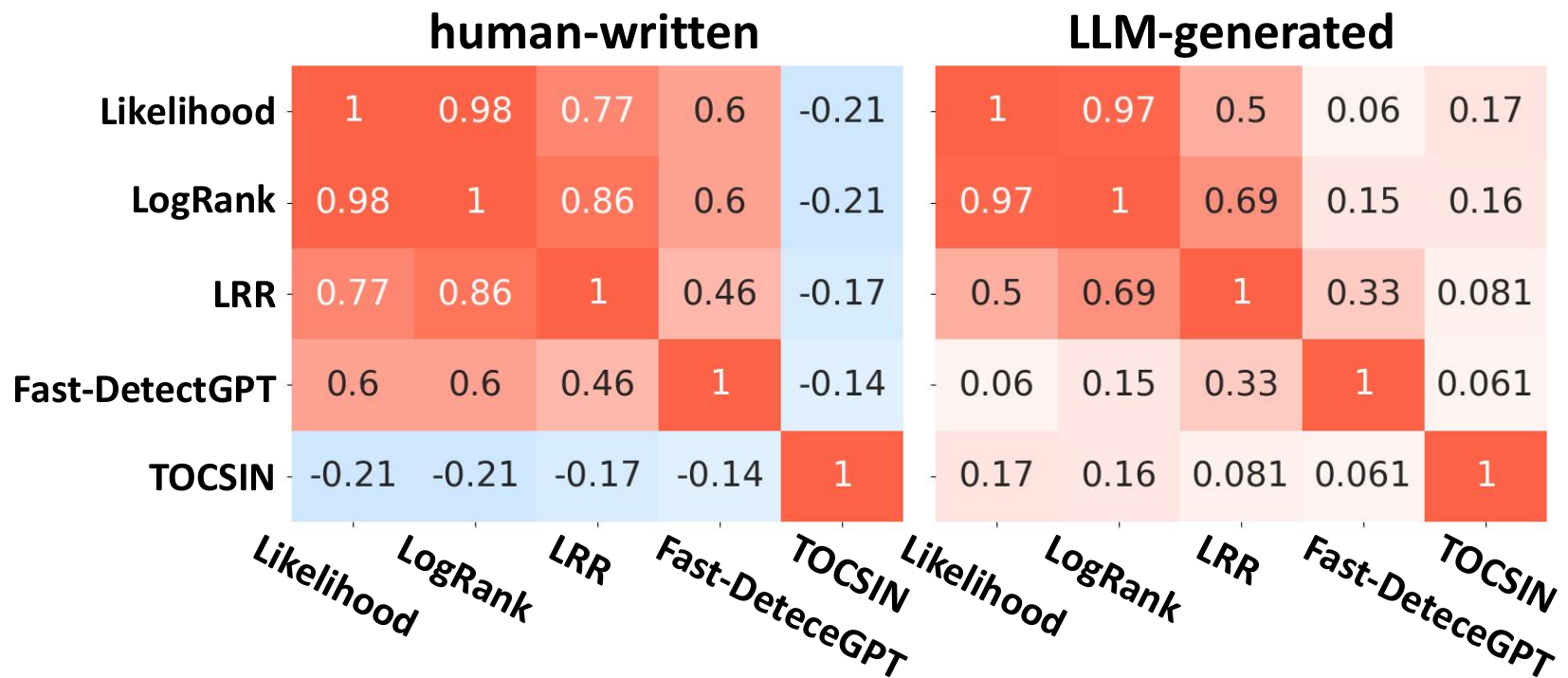}
  \caption{Heatmaps of Pearson Correlation Coefficient between scores from different detectors, averaged across XSum, SQuAD, WritingPrompts and five open-source models. Lighter colors indicate lower correlation, while darker colors indicate stronger correlation.}
  \label{fig:heatmap-correlation}
\end{figure}

\paragraph{Complementarity with Existing Detectors}
The success of TOCSIN is largely attributed to the good complementarity between the token cohesiveness feature and existing detectors. To validate this viewpoint, we compare the performance of combining Fast-DetectGPT with an existing detector versus combining it with token cohesiveness, in the same manner as described in Section~\ref{subsec:tocsin}. Table~\ref{tab:Ensemble} reports average AUROC across XSum, SQuAD, WritingPrompts for the five open-source models, showing that combining Fast-DetectGPT with Likelihood, LogRank, or LRR does not always yield substantial improvements as it does with TOCSIN.

We further examine the Pearson Correlation Coefficient between the Likelihood, LogRank, LRR, Fast-DetectGPT, and token cohesiveness scores for the LLM-generated and human-written passages therein. Figure \ref{fig:heatmap-correlation} visualizes the results, averaged across the three datasets and five source models. From the figure, we can observe a relatively high positive correlation among existing detectors, particularly for human-written text, whereas their correlation with the token cohesiveness scores is rather low. This observation indicates strong complementarity between token cohesiveness and existing detectors, which we think is key to the success of our dual-channel detection paradigm. Note that similar to Likelihood, LogRank, LRR and Fast-DetectGPT, token cohesiveness can also be used alone for zero-shot detection of LLM-generated text, the performance of which is shown in Appendix~\ref{subsec:standalone}.

\begin{figure}[t]
  \centering
  \includegraphics[width=0.48\textwidth]{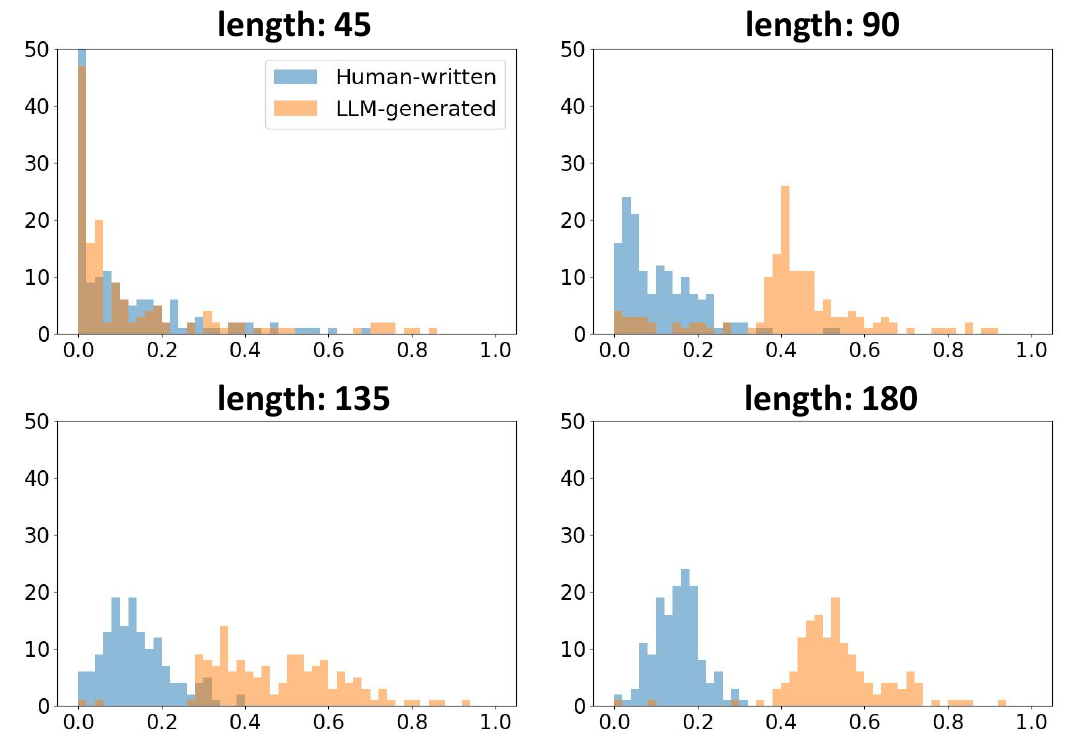}
  \caption{Distribution of token cohesiveness between 150 human-written and 150 ChatGPT-generated passages from WritingPrompts truncated to target length.}
  \label{fig:cohesiveness-length}
\end{figure}

\begin{figure}[t]
  \centering
  \includegraphics[width=0.48\textwidth]{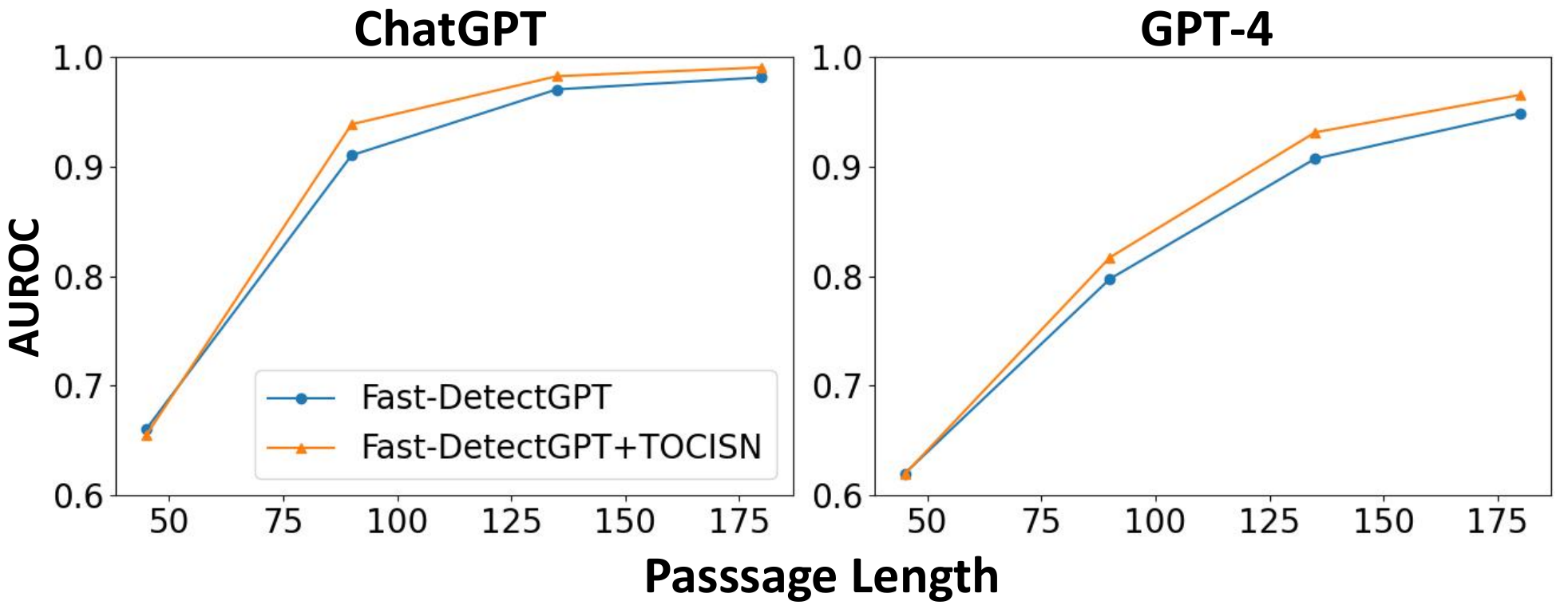}
  \caption{AUROC for detecting ChatGPT and GPT-4 passages on WritingPrompts truncated to target length.}
  \label{fig:performance-length}
\end{figure}

\paragraph{Impact of Passage Length}
We have observed in Table~\ref{tab:api-based-models} that TOCSIN may not perform that well on shorter passages. To rigorously evaluate the impact of passage length, we truncate the passages from WritingPrompts to various target lengths of 45, 90, 135, 180, and explore how the token cohesiveness and overall performance of TOCSIN varies at the four different lengths. The results are reported in Figure~\ref{fig:cohesiveness-length} and Figure~\ref{fig:performance-length}. We can see that at a shorter length of 45, the distributions of token cohesiveness between human-written and LLM-generated text overlap significantly and cannot be discriminated. TOCSIN also fails to improve FastDetect-GPT at this passage length. But as the length increases, the disparities in token cohesiveness between the two types of text become more obvious, and incorporating token cohesiveness into FastDetect-GPT starts to achieve consistent improvements. These results suggest that TOCSIN is more suitable for detecting long passages generated by LLMs.

\begin{figure*}[t]
  \centering
  \includegraphics[width=1\textwidth]{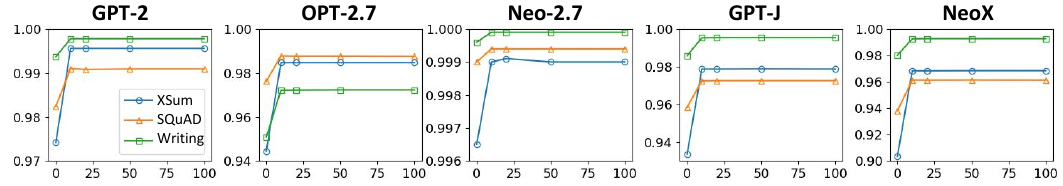}\\
  \includegraphics[width=1\textwidth]{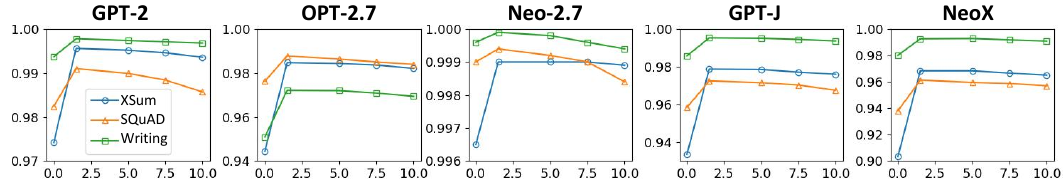}
  \caption{AUROC of FastDetectGPT+TOCSIN with varying hyperparameters on XSum, SQuAD, and WritingPrompts across five open-source LLMs in the black-box setting. {\bf Top:} number of copies $n \in \{10, 20, 50, 100\}$ and token deletion proportion $\rho=1.5\%$ where $n=0$ denotes AUROC of Fast-DetectGPT. {\bf Bottom:} $\rho \in \{1.5\%, 5.0\%,$ $7.5\%, 10.0\%\}$ and $n=10$ where $\rho=0.0\%$ denotes AUROC of Fast-DetectGPT.}
  \label{fig:hyperparameters}
\end{figure*}

\paragraph{Impact of Hyperparameters}
TOCSIN gets two hyperparameters: the number of copies created for each input ($n$) and the proportion of tokens deleted in each copy ($\rho$). We examine the impact of the two hyperparameters, and show how the performance of Fast-DetectGPT+TOCSIN varies as $n \in \{10,$ $20, 50, 100\}$ and $\rho \in \{1.5\%, 5.0\%, 7.5\%, 10.0\%\}$. Figure~\ref{fig:hyperparameters} presents the results on XSum, SQuAD, and WritingPrompts with LLM-generated passages from the five open-source models, where $n=0$ and $\rho=0.0\%$ denote the performance of the base detector Fast-DetectGPT. The results suggest that the performance of TOCSIN is not sensitive to the hyperparameters. TOCSIN performs rather stably with different $n$ values, so we just use $n=10$ for simplicity and efficiency. Moreover, a smaller $\rho$ value of $1.5\%$ generally performs better, and the performance drops slightly as $\rho$ grows up.

\section{Conclusion}
This paper introduces the concept of token cohesiveness and unveils that it can serve as a new criterion to discriminate LLM-generated and human-written text. Based on this new finding, we devise TOCSIN for zero-shot detection of LLM-generated text. TOCSIN is a generic dual-channel detection paradigm that uses token cohesiveness as a plug-and-play module to improve existing zero-shot detectors. As empirical evaluation, we apply TOCSIN to four current state-of-the-art base detectors, and achieve meaningful improvements across four diversified datasets with passages generated from eight different source LLMs, demonstrating the effectiveness and generality of our approach.

\section*{Limitations}
This work has two limitations. First, the proposed method TOCSIN, like most zero-shot detectors, is more suitable for long text and has limited effectiveness on short text. For example, it consistently performs well on passages consisting of 90 tokens or more, but fails on short passages with only 45 tokens, as illustrated in Figure~\ref{fig:performance-length}. Second, this work is restricted to the most basic form of LLM-generated text detection, i.e., binary classification of LLM-generated and human-written text. Whether TOCSIN still works in more challenging tasks, such as detecting mixtures of LLM-generated and human-written text and tracing the origin of generation, remains an open question for future research.

\section*{Acknowledgements}
We would like to thank Xingyu Yao for preparing the Gemini datasets. We would also like to thank the action editor and the reviewers for their insightful and valuable suggestions, which significantly improve the quality of this work. This work is supported by the National Natural Science Foundation of China (grants No. 62376033 and 62232006). 

\bibliography{custom}

\appendix
\section{Semantic Difference Metrics}\label{appendix:similarity-metrics}
We use the negative BARTScore~\cite{NEURIPS2021_e4d2b6e6} to measure the semantic difference between input $x$ and its copy $\tilde{x}$ with some tokens randomly deleted in our main experiments, i.e.,
\begin{equation*}
\mathrm{DIFF}(x,\tilde{x}) = - \mathrm{BARTScore}(x,\tilde{x}).
\end{equation*}
Besides, we also evaluate another semantic difference metric, negative GPTScore~\cite{fu-etal-2024-gptscore}, and discuss the results in Appendix~\ref{subsec:gptscore}.
\begin{equation*}
\mathrm{DIFF}(x,\tilde{x}) = - \mathrm{GPTScore}(x,\tilde{x}).
\end{equation*}
Below we introduce the two metrics in detail.

\paragraph{BARTScore}
This metric, built on a BART model \cite{lewis-etal-2020-bart} parameterized by $\phi$, is calculated as the log probability of generating the input $x$ (target text) conditioned on its copy $\tilde{x}$ (source text), which can be factorized as:
\begin{equation*}
\mathrm{BARTScore}(x,\tilde{x}) = \sum_{j=1}^k \log p_\phi (x_j|x_{<j}, \tilde{x}).
\end{equation*}
Here, $k$ is the total number of tokens in $x$, $x_j$ the $j$-th token therein, and $x_{<j}$ the sequence preceding $x_j$. This score essentially measures the semantic coverage between the source and target text, and its negative value therefore measures their semantic difference. In this paper, we use BART-base which has 139M parameters to compute BARTScore, so as to ensure the high efficiency of token cohesiveness calculation. Note that this model is typically much smaller than the scoring model required by the base detector, i.e., the source model itself in the white-box setting and the surrogate model (e.g., the 2.7B GPT-Neo-2.7) in the black-box setting. 

\paragraph{GPTScore}
This metric is similar to BARTScore but has two differences. First, it is built on a GPT model \cite{radford2019language}, parameterized by $\phi$, rather than BART. Furthermore, it considers the generation of the target text (input $x$) conditioned on a more complex source text $T(x,\tilde{x})$, rather than just on $\tilde{x}$ itself. $T(x,\tilde{x})$ is specified as ``$\tilde{x}\; [\textrm{In other}$ $\textrm{words,}] \;x$'' in the semantic similarity measurement protocol. Summarizing the above two differences, GPTScore is formally defined as:
\begin{equation*}
\mathrm{GPTScore}(x,\tilde{x}) = \sum_{j=1}^k \log p_\phi (x_j|x_{<j}, T(x,\tilde{x})).
\end{equation*}
We use GPT-2-small with 117M parameters to compute GPTScore, which is also much smaller than the scoring model required by the base detector.

\section{Base Detectors}\label{appendix:base-detector}
In this paper we choose Likelihood \cite{mitchell2023detectgpt}, logRank \cite{mitchell2023detectgpt}, LRR \cite{su2023detectllm} and Fast-DetectGPT \cite{bao2024fastdetectgpt}, which are computationally efficient and report current state-of-the-art performance as the base detectors. Each base detector requires a scoring model $\theta$ to score passages, which is the source LLM in the white-box setting and the surrogate model in the black-box setting (GPT-Neo-2.7 in this paper). Below we introduce the four base detectors in detail.

\paragraph{Likelihood}
Given a candidate passage $x$, Likelihood is formally defined as:
\begin{equation*}
\mathrm{Likelihood}(x) = {\sum_{j=1}^k \log p_\theta (x_j | x_{<j})} ,
\end{equation*}
where $p_\theta (x_j | x_{<j})$ is the probability of token $x_j$ conditioned on its preceding tokens predicted by the scoring model $\theta$. LLM-generated passages are supposed to have higher Likelihood scores compared to human-written passages.

\paragraph{LogRank}
Given a candidate passage $x$, LogRank is formally defined as:
\begin{equation*}
\mathrm{LogRank}(x) = -{\sum_{j=1}^k \log r_\theta (x_j | x_{<j})} ,
\end{equation*}
where $r_\theta (x_j | x_{<j})$ denotes the rank of the probability of token $x_j$ conditioned on its preceding tokens predicted by the scoring model $\theta$. LLM-generated text tends to have higher LogRank scores compared to human-written text.

\paragraph{LRR}
LRR makes use of the Log-Likelihood Log-Rank Ratio to discern LLM-generated and human-written text. Given a candidate passage $x$, LRR is formally defined as:
\begin{equation*}
\mathrm{LRR}(x) = - \frac{\sum_{j=1}^k \log p_\theta (x_j | x_{<j})} {\sum_{j=1}^k \log r_\theta (x_j | x_{<j})},
\end{equation*}
which can be seen as a combination of Likelihood and LogRank. LLM-generated text is supposed to have higher LRR scores than human-written text.

\paragraph{Fast-DetectGPT}
Fast-DetectGPT makes use of a conditional probability function to detect LLM-generated text. Given a passage $x$, it first performs conditional independent sampling from $q_\psi(\cdot|x)$ to create a group of samples $\{\tilde{x}_1, \tilde{x}_2, \cdots, \tilde{x}_n\}$. This sampling samples alternative word choices at each token conditioned on the fixed passage $x$ without depending on other sampled tokens. Then, it evaluates the conditional probabilities $p_\theta (\tilde{x}|x)$ of these samples and combines them to arrive at a decision:
\begin{equation*}
\mathrm{Fast}\textrm{-}\mathrm{DetectGPT}(x) = \frac{1}{n} \sum_{i=1}^n \log \frac{p_\theta (x|x)}{p_\theta (\tilde{x}|x)}.
\end{equation*}
If the score exceeds a specific threshold, the passage is probably LLM-generated. Fast-DetectGPT invokes the sampling model $q_\psi(\cdot|x)$ once to generate all samples and similarly the scoring model $p_\theta(\cdot|x)$ once to evaluate all samples, and therefore is rather efficient. In the white-box setting, $q_\psi(\cdot|x)$ and $p_\theta(\cdot|x)$ are both set to the source LLM. In the black-box setting, $q_\psi(\cdot|x)$ is set to the 6B GPT-J and $p_\theta(\cdot|x)$ the 2.7B GPT-Neo-2.7. All configurations are identical to those in \cite{bao2024fastdetectgpt}.

\begin{table*}[!t]
  \centering\small
  \begin{tabular*}{1 \textwidth}{@{\extracolsep{\fill}}llcccccc}
    \toprule
    {\bf Dataset} & {\bf Method} & {\bf GPT-2} & {\bf OPT-2.7} & {\bf Neo-2.7} & {\bf GPT-J} & {\bf NeoX} & {\bf Avg.} \\
    \midrule
    \multirow{16}{*}{XSum}
    &Entropy \cite{mitchell2023detectgpt}         &0.5835 &0.5071 &0.5712  &0.5705 &0.6035 &0.5671 \\
    &DNA-GPT \cite{yang2024dnagpt}$\dag$ &0.8548  &0.8168  &0.8197 &0.7586  &0.7167  &0.7933  \\ 
    &DetectGPT \cite{mitchell2023detectgpt}$\dag$ &0.9875  &0.9621  &0.9914  &0.9632  &0.9398  &0.9688  \\    
    &NPR \cite{su2023detectllm}$\dag$     &0.9891  &0.9681  &0.9929  &0.9566  &0.9311  &0.9676  \\
    \cdashline{2-8}
    &Likelihood \cite{mitchell2023detectgpt}     &0.8638 &0.8600 &0.8609  &0.8101 &0.7604 &0.8310  \\  
    &Likelihood+TOCSIN (ours)       & {\bf0.9943 } & {\bf0.9917 } & {\bf 0.9895} & {\bf 0.9864} & {\bf0.9829 } & {\bf 0.9890} \\
    &{\it(Absolute $\uparrow$)} & {\it 13.05\%}  & {\it 13.17\%}  & {\it 12.86\%}  & {\it 17.63\%} & {\it 22.25\%} & {\it 15.80\%} \\
    \cdashline{2-8}
    &LogRank \cite{mitchell2023detectgpt}       &0.8918 &0.8839 &0.8949  &0.8407 &0.7939 & 0.8610 \\ 
    &LogRank+TOCSIN (ours)       & {\bf0.9950 } & {\bf 0.9932} & {\bf0.9916 } & {\bf 0.9877} & {\bf 0.9838} & {\bf 0.9903} \\
    &{\it(Absolute $\uparrow$)} & {\it 10.32\%}  & {\it 10.93\%}  & {\it9.67\%}  & {\it 14.70\%} & {\it 18.99\%} & {\it 12.93\%} \\
    \cdashline{2-8}
    &LRR \cite{su2023detectllm}  &0.9179  &0.8867  &0.9190  &0.8592  &0.8205  &0.8807  \\ 
    &LRR+TOCSIN (ours)       & {\bf 0.9957} & {\bf 0.9968} & {\bf 0.9935} & {\bf 0.9950} & {\bf 0.9926} & {\bf 0.9947} \\
    &{\it(Absolute $\uparrow$)} & {\it 7.78\%}  & {\it 11.01\%}  & {\it 7.45\%}  & {\it 13.58\%} & {\it 17.21\%} & {\it 11.40\%} \\
    \cdashline{2-8}
    &Fast-DetectGPT \cite{bao2024fastdetectgpt}  &0.9930  &0.9803  &0.9885  &0.9771  & 0.9703 & 0.9818 \\ 
    &Fast-DetectGPT+TOCSIN (ours) & {\bf 0.9974} & {\bf 0.9928} & {\bf 0.9966} & {\bf 0.9927} & {\bf 0.9850} & {\bf 0.9929} \\
    &{\it(Absolute $\uparrow$)} & {\it 0.44\%} & {\it 1.25\%} & {\it 0.81\%} & {\it 1.56\%} & {\it 1.47\%} & {\it 1.11\%} \\
    \midrule
    \multirow{16}{*}{SQuAD}
    &Entropy \cite{mitchell2023detectgpt}         &0.5791  & 0.5119 &0.5581  & 0.5643 & 0.6056 &0.5638  \\
    &DNA-GPT \cite{yang2024dnagpt}$\dag$ &0.9094  &0.8934  &0.8589 &0.8069  & 0.7525 & 0.8442 \\ 
    &DetectGPT \cite{mitchell2023detectgpt}$\dag$ &0.9914  &0.9763  & 0.9625 & 0.8738 &0.7916  &0.9191  \\    
    &NPR \cite{su2023detectllm}$\dag$                    & 0.9965 &0.9853  &0.9789  &0.9108  &0.8175  &0.9378  \\
    \cdashline{2-8}
    &Likelihood \cite{mitchell2023detectgpt}     & 0.9077 & 0.8839 &0.8585  &0.7943  &0.6977  &0.8284  \\  
    &Likelihood+TOCSIN (ours)       & {\bf 0.9888} & {\bf 0.9733} & {\bf 0.9608} & {\bf0.9562 } & {\bf0.8935 } & {\bf 0.9545} \\
    &{\it(Absolute $\uparrow$)} & {\it 8.11\%}  & {\it 8.94\%}  & {\it 10.23\%}  & {\it 16.19\%} & {\it 19.58\%} & {\it 12.61\%} \\
    \cdashline{2-8}
    &LogRank \cite{mitchell2023detectgpt}       &0.9454  &0.9203  &0.9054  &0.8471  &0.7545  & 0.8745 \\ 
    &LogRank+TOCSIN (ours)       & {\bf0.9928 } & {\bf 0.9814} & {\bf 0.9739} & {\bf 0.9630} & {\bf 0.9019} & {\bf 0.9626} \\
    &{\it(Absolute $\uparrow$)} & {\it 4.74\%}  & {\it 6.11\%}  & {\it 6.85\%}  & {\it 11.59\%} & {\it 14.74\%} & {\it 8.81\%} \\
    \cdashline{2-8}
    &LRR \cite{su2023detectllm}  &0.9773  & 0.9597 &0.9610  &0.9244  &0.8600  &0.9365  \\ 
    &LRR+TOCSIN (ours)       & {\bf 0.9928} & {\bf 0.9836} & {\bf 0.9777} & {\bf 0.9862} & {\bf 0.9621} & {\bf 0.9805} \\
    &{\it(Absolute $\uparrow$)} & {\it 1.55\%}  & {\it 2.39\%}  & {\it 1.67\%}  & {\it 6.18\%} & {\it 10.21\%} & {\it 4.40\%} \\
    \cdashline{2-8}
    &Fast-DetectGPT \cite{bao2024fastdetectgpt}  &0.9990  &0.9949  & 0.9956 &0.9853  & 0.9617 &0.9873  \\ 
    &Fast-DetectGPT+TOCSIN (ours) & {\bf 0.9996} & {\bf 0.9972} & {\bf 0.9975} & {\bf 0.9904} & {\bf 0.9764} & {\bf 0.9922} \\
    &{\it(Absolute $\uparrow$)} & {\it 0.06\%} & {\it 0.23\%} & {\it 0.19\%} & {\it 0.51\%} & {\it 1.47\%} & {\it 0.49\%} \\
    \midrule
    \multirow{16}{*}{WritingPrompts} 
    &Entropy \cite{mitchell2023detectgpt}         &0.3895  & 0.4299 &0.3400  &0.3668  & 0.3908 &0.3834  \\
    &DNA-GPT \cite{yang2024dnagpt}$\dag$ &0.9431  & 0.9288 &0.9283  &0.9026 &0.8786 &0.9163  \\ 
    &DetectGPT \cite{mitchell2023detectgpt}$\dag$ &0.9962  &0.9891  & 0.9852 & 0.9688 & 0.9516 &0.9782  \\    
    &NPR \cite{su2023detectllm}$\dag$        &0.9987  &0.9962  &0.9930  &0.9825  &0.9708  &0.9882  \\
    \cdashline{2-8}
    &Likelihood \cite{mitchell2023detectgpt}     &0.9661  & 0.9451 &0.9505  &0.9396 &0.9256  & 0.9454 \\  
    &Likelihood+TOCSIN (ours)       & {\bf0.9884 } & {\bf0.9976 } & {\bf 0.9880} & {\bf 0.9901} & {\bf0.9882 } & {\bf 0.9905} \\
    &{\it(Absolute $\uparrow$)} & {\it 2.23\%}  & {\it 5.25\%}  & {\it 3.75\%}  & {\it 5.05\%} & {\it 6.26\%} & {\it 4.51\%} \\
     \cdashline{2-8}
    &LogRank \cite{mitchell2023detectgpt}       &0.9782  &0.9628 &0.9675  & 0.9577 & 0.9454 & 0.9623 \\ 
    &LogRank+TOCSIN (ours)       & {\bf0.9922 } & {\bf 0.9976} & {\bf 0.9916} & {\bf0.9927 } & {\bf 0.9902} & {\bf 0.9929} \\
    &{\it(Absolute $\uparrow$)} & {\it 1.40\%}  & {\it 3.48\%}  & {\it 2.41\%}  & {\it 3.50\%} & {\it 4.48\%} & {\it 3.06\%} \\
    \cdashline{2-8}
    &LRR \cite{su2023detectllm}  & 0.9850 & 0.9740 &0.9766  &0.9702  &0.9573  & 0.9726 \\ 
    &LRR+TOCSIN (ours)       & {\bf 0.9871} & {\bf 0.9983} & {\bf 0.9907} & {\bf 0.9929} & {\bf 0.9850} & {\bf 0.9908} \\
    &{\it(Absolute $\uparrow$)} & {\it 0.21\%}  & {\it 2.43\%}  & {\it 1.41\%}  & {\it 2.27\%} & {\it 2.77\%} & {\it 1.82\%} \\
    \cdashline{2-8}
    &Fast-DetectGPT \cite{bao2024fastdetectgpt}  & 0.9982 & 0.9972 & 0.9980 & 0.9974 & 0.9941 &0.9970  \\ 
    &Fast-DetectGPT+TOCSIN (ours) & {\bf 0.9988} & {\bf 0.9979} & {\bf 0.9993} & {\bf 0.9992} & {\bf 0.9974} & {\bf 0.9985} \\
    &{\it(Absolute $\uparrow$)} & {\it 0.06\%} & {\it 0.07\%} & {\it 0.13\%} & {\it 0.18\%} & {\it 0.33\%} & {\it 0.15\%} \\
    \bottomrule
  \end{tabular*}
  \caption{Details of the main results in Table~\ref{tab:open-source-models} on three datasets in {\bf white-box setting}, with all setups identical to those in Table~\ref{tab:open-source-models}. 
 }
  \label{tab:open-source-models_white-box}
\end{table*}

\begin{table*}[!t]
  \centering\small
  \begin{tabular*}{1 \textwidth}{@{\extracolsep{\fill}}llcccccc}
    \toprule
    {\bf Dataset} & {\bf Method} & {\bf GPT-2} & {\bf OPT-2.7} & {\bf Neo-2.7} & {\bf GPT-J} & {\bf NeoX} & {\bf Avg.} \\
    \midrule
    \multirow{13}{*}{XSum} 
    &DetectGPT \cite{mitchell2023detectgpt}$\dag$ &0.9180  &0.8868  &0.9914  &0.8830  &0.8682  &0.9095  \\
    \cdashline{2-8}
    &Likelihood (ours) &  0.7308& 0.7918 & 0.8609 & 0.7508 &0.7429 & 0.7754 \\
    &Likelihood+TOCSIN (ours) & {\bf 0.9901} & {\bf0.9891 } & {\bf 0.9895} & {\bf0.9829 } & {\bf0.9825 } & {\bf 0.9868} \\
    &{\it(Absolute $\uparrow$)} & {\it 25.93\%} & {\it 19.73\%} & {\it 12.86\%} & {\it 23.21\%} & {\it 23.96\%} & {\it 21.14\%} \\
    \cdashline{2-8}
    &LogRank (ours) & 0.7610 & 0.8139 & 0.8950 &0.7747  &0.7550  &0.7999  \\
    &LogRank+TOCSIN (ours) & {\bf0.9887} & {\bf 0.9901} & {\bf0.9916 } & {\bf0.9824 } & {\bf0.9807} & {\bf 0.9867} \\
    &{\it(Absolute $\uparrow$)} & {\it 22.77\%} & {\it 17.62\%} & {\it 9.66\%} & {\it 20.77\%} & {\it 22.57\%} & {\it 18.68\%} \\
    \cdashline{2-8}
    &LRR (ours) & 0.7824 & 0.8069 &0.9186  & 0.7767 & 0.7357 &0.8041  \\
    &LRR+TOCSIN (ours) & {\bf 0.9904} & {\bf 0.9953} & {\bf 0.9935} & {\bf 0.9914} & {\bf 0.9901} & {\bf 0.9921} \\
    &{\it(Absolute $\uparrow$)} & {\it 20.80\%} & {\it 18.84\%} & {\it 7.49\%} & {\it 21.47\%} & {\it 25.44\%} & {\it 18.80\%} \\
    \cdashline{2-8}
    &Fast-DetectGPT \cite{bao2024fastdetectgpt} &0.9742  &0.9444  &0.9965 &0.9335  &0.9033  &0.9504  \\
    &Fast-DetectGPT+TOCSIN (ours) & {\bf 0.9956} & {\bf 0.9847} & {\bf 0.9990} & {\bf 0.9787} & {\bf 0.9683} & {\bf 0.9853} \\
    &{\it(Absolute $\uparrow$)} & {\it 2.14\%} & {\it 4.03\%} & {\it 0.25\%} & {\it 4.52\%} & {\it 6.50\%} & {\it 3.49\%} \\
    \midrule
    \multirow{13}{*}{SQuAD} 
    &DetectGPT \cite{mitchell2023detectgpt}$\dag$ &0.7382  &0.7530  &0.9625  &0.7882  & 0.7709 & 0.8026 \\
    \cdashline{2-8}
    &Likelihood (ours) & 0.6772 &0.7372 &0.8584 &0.7562 & 0.7206&0.7499  \\
    &Likelihood+TOCSIN (ours) & {\bf 0.9382} & {\bf0.9346 } & {\bf0.9608 } & {\bf0.9480 } & {\bf 0.9229} & {\bf 0.9409} \\
    &{\it(Absolute $\uparrow$)} & {\it 26.10\%} & {\it 19.74\%} & {\it 10.24\%} & {\it 19.18\%} & {\it 20.23\%} & {\it 19.10\%} \\
    \cdashline{2-8}
    &LogRank (ours) & 0.7387 &  0.7877& 0.9052 & 0.8042 &  0.7579&0.7987  \\
    &LogRank+TOCSIN (ours) & {\bf 0.9378} & {\bf0.9437 } & {\bf 0.9739} & {\bf 0.9512} & {\bf 0.9243} & {\bf 0.9462} \\
    &{\it(Absolute $\uparrow$)} & {\it 19.91\%} & {\it 15.60\%} & {\it 6.87\%} & {\it 14.7\%} & {\it 16.64\%} & {\it 14.75\%} \\
    \cdashline{2-8}
    &LRR (ours) &0.8447  &0.8615  & 0.9603 & 0.8709 &  0.8109&0.8697  \\
    &LRR+TOCSIN (ours) & {\bf 0.9713} & {\bf 0.9620} & {\bf 0.9777} & {\bf 0.9701} & {\bf 0.9552} & {\bf 0.9673} \\
    &{\it(Absolute $\uparrow$)} & {\it 12.66\%} & {\it 10.05\%} & {\it 1.74\%} & {\it 9.92\%} & {\it 14.43\%} & {\it 9.76\%} \\
    \cdashline{2-8}
    &Fast-DetectGPT \cite{bao2024fastdetectgpt} & 0.9824 & 0.9762 &0.9990&0.9584  &0.9379  & 0.9708 \\
    &Fast-DetectGPT+TOCSIN (ours) & {\bf 0.9910} & {\bf 0.9878} & {\bf 0.9994} & {\bf 0.9725} & {\bf 0.9613} & {\bf 0.9824} \\
    &{\it(Absolute $\uparrow$)} & {\it 0.86\%} & {\it 1.16\%} & {\it 0.04\%} & {\it 1.41\%} & {\it 2.34\%} & {\it 1.16\%} \\
    \midrule
    \multirow{13}{*}{WritingPrompts} 
    &DetectGPT \cite{mitchell2023detectgpt}$\dag$ & 0.8989 & 0.8772 &0.9852  & 0.9014 & 0.8809 &0.9087  \\
    \cdashline{2-8}
    &Likelihood (ours) &  0.8795&0.8225  & 0.9505 & 0.9093 &0.8919  & 0.8907\\
    &Likelihood+TOCSIN (ours) & {\bf0.9596 } & {\bf0.9933 } & {\bf 0.9880} & {\bf 0.9857} & {\bf0.9831 } & {\bf 0.9819} \\
    &{\it(Absolute $\uparrow$)} & {\it 8.01\%} & {\it 17.08\%} & {\it 3.75\%} & {\it 7.64\%} & {\it 9.12\%} & {\it 9.12\%} \\
    \cdashline{2-8}
    &LogRank (ours) &  0.9043& 0.8614 &0.9675  & 0.9298 & 0.9081 &0.9142 \\
    &LogRank+TOCSIN (ours) & {\bf 0.9667} & {\bf0.9922 } & {\bf 0.9916 } & {\bf0.9874 } & {\bf0.9839 } & {\bf 0.9844} \\
    &{\it(Absolute $\uparrow$)} & {\it 6.24\%} & {\it 13.08\%} & {\it 2.41\%} & {\it 5.76\%} & {\it 7.58\%} & {\it 7.02\%} \\
    \cdashline{2-8}
    &LRR (ours) &0.9244  & 0.9142 &0.9766  &0.9436  & 0.9095 & 0.9337\\
    &LRR+TOCSIN (ours) & {\bf 0.9617} & {\bf 0.9975} & {\bf 0.9907} & {\bf 0.9880} & {\bf 0.9803} & {\bf 0.9836} \\
    &{\it(Absolute $\uparrow$)} & {\it 3.73\%} & {\it 8.33\%} & {\it 1.41\%} & {\it 4.44\%} & {\it 7.08\%} & {\it 4.99\%} \\
    \cdashline{2-8}
    &Fast-DetectGPT \cite{bao2024fastdetectgpt} &0.9937  &0.9509  & 0.9996 &0.9858  & 0.9801 &0.9820  \\
    &Fast-DetectGPT+TOCSIN (ours) & {\bf 0.9978} & {\bf 0.9721} & {\bf 0.9999} & {\bf 0.9953} & {\bf 0.9926} & {\bf 0.9915} \\
    &{\it(Absolute $\uparrow$)} & {\it 0.41\%} & {\it 2.12\%} & {\it 0.03\%} & {\it 0.95\%} & {\it 1.25\%} & {\it 0.95\%} \\
    \bottomrule
  \end{tabular*}
  \caption{Details of the main results in Table~\ref{tab:open-source-models} on three datasets in {\bf black-box setting}, with all setups identical to those in Table~\ref{tab:open-source-models}. 
 }
  \label{tab:open-source-models_black-box}
\end{table*}

\section{Datasets Created by Gemini}\label{appendix:Gemini}
We follow exactly the same procedure as in \cite{bao2024fastdetectgpt} to generate samples for XSum, WritingPrompts, and PubMedQA by calling the Gemini API. Specifically, we request chat completions with predefined instructions as follows.

\begin{figure}[h]
\begin{center}
{\small\fbox{
\begin{varwidth}{0.97 \columnwidth}
{\bf Instruction for XSum} \\
\texttt{\{`system': `You are a News writer.'\} \\
\{`user': `Please write an article with about 150 words starting exactly with: <prefix>'\}}
\vspace{-5pt}\\\rule{\linewidth}{0.1mm}
{\bf Instruction for WritingPrompts} \\
\texttt{\{`system': `You are a Fiction writer.'\} \\
\{`user': `Please write an article with about 150 words starting exactly with: <prefix>'\}}
\vspace{-5pt}\\\rule{\linewidth}{0.1mm}
{\bf Instruction for PubMedQA} \\
\texttt{\{`system': `You are a Technical writer.'\} \\
\{`user': `Please answer the question in about 50 words. <prefix>'\}}
\end{varwidth}}}
\end{center}
\end{figure}

\noindent Here \texttt{<prefix>} is a prefix consisting of the initial 30 tokens of a human-written passage, e.g., ``Maj Richard Scott, 40, is accused of driving at speeds of up to 95mph (153km/h) in bad weather'', and the response is supposed to start with it. We suppose to create 150 samples for each of the three datasets, but unfortunately fail on 19 samples for XSum and 7 for WritingPrompts, resulting in a total of 131 and 143 samples for these two datasets, respectively.

\section{Additional Experimental Results}\label{appendix:additional-exp}
\subsection{Detailed Results of Open-Source LLMs}\label{subsec:open-source-models}
Table~\ref{tab:open-source-models_white-box} presents AUROC for zero-shot detection of passages generated by the five open-source LLMs on three datasets of XSum, SQuAD, and WritingPrompts in the white-box setting, and Table~\ref{tab:open-source-models_black-box} reports the same results in the black-box setting. As we can see, TOCSIN brings consistent improvements in both settings, regardless of the datasets, source models, or base detectors.

\begin{table*}[!t]
  \centering\small\setlength{\tabcolsep}{4pt}
  \begin{tabular*}{1 \textwidth}{@{\extracolsep{\fill}}llcccccccccc}
    \toprule
    \multirow{2}{*}{\bf Dataset} & \multirow{2}{*}{\bf Method}
    &\multicolumn{5}{c}{\textbf{Runtime (s)}} & \multicolumn{5}{c}{\textbf{GPU Memory (GB)}}\\
    \cmidrule(r){3-7}\cmidrule(l){8-12}
    &&{\bf GPT-2} &{\bf OPT-2.7} & {\bf Neo-2.7} &{\bf GPT-J} &{\bf NeoX} 
    &{\bf GPT-2} &{\bf OPT-2.7} & {\bf Neo-2.7} &{\bf GPT-J} &{\bf NeoX} \\
    \midrule
    \multirow{3}{*}{XSum} &
    w/o TOCSIN & 0.31&0.31 & 0.31& 0.31&0.31 & 23.57&23.30 &23.22 &24.21 &23.53 \\
    & w/ TOCSIN & 0.48& 0.46&0.46 & 0.47&0.46 &28.35 & 27.80& 28.37& 28.79&27.35 \\
    & {\it(Absolute $\uparrow$)} &{\it 0.17} &{\it 0.15} &{\it 0.15} &{\it 0.16} & {\it 0.15}& {\it ~~4.78}& {\it ~~4.50}& {\it ~~5.15}& {\it ~~4.58}&{\it ~~3.82}\\
    \midrule
    \multirow{3}{*}{SQuAD} &
    w/o TOCSIN &0.32 &0.32&0.32 &0.32 &0.31& 23.35& 23.98&23.90 & 24.05&24.00\\
    & w/ TOCSIN &0.49 &0.46 &0.49 &0.48 & 0.47& 28.14&28.53 &28.54 &28.07 &29.19 \\
    & {\it(Absolute $\uparrow$)}& {\it 0.17}& {\it 0.14}& {\it 0.17}& {\it 0.16}&{\it 0.16} & {\it ~~4.79}& {\it ~~4.55}& {\it ~~4.64}&{\it ~~4.02} &{\it ~~5.19} \\
    \midrule
    \multirow{3}{*}{Writing} &
    w/o TOCSIN &0.32 &0.30 &0.31 &0.31 &0.31 &24.98 &24.30 &23.89 &24.75 & 24.35\\
    & w/ TOCSIN &0.48 & 0.45& 0.48& 0.47& 0.47&29.31 & 29.24&29.51 &29.56 &29.32  \\
    & {\it(Absolute $\uparrow$)}& {\it 0.16}& {\it 0.15}& {\it 0.17}&{\it 0.16} &{\it 0.16} & {\it ~~4.33}& {\it ~~4.94}&{\it ~~5.62} &{\it ~~4.81} &{\it ~~4.97} \\
    \bottomrule
  \end{tabular*}
  \caption{
   Runtime per instance and GPU memory usage of w/ and w/o TOCSIN variants of Fast-DetectGPT in black-box setting. ``{\it(Absolute $\uparrow$)}'' means additional time/space cost brought by TOCSIN.
  }
  \label{tab:time-space_details}
\end{table*}

\begin{table*}[t]
  \centering\small
  \begin{tabular*}{1 \textwidth}{@{\extracolsep{\fill}}lcccccc}
    \toprule
    {\bf Method} & {\bf GPT-2} & {\bf OPT-2.7} & {\bf Neo-2.7} & {\bf GPT-J} & {\bf NeoX} & {\bf Avg.} \\
    \midrule
    Likelihood  & 0.7625&0.7838  & 0.8899& 0.8054 &0.7851 & 0.8053\\  
    LogRank & 0.8013& 0.8210 &0.9226 &0.8362  & 0.8070&0.8376 \\ 
    LRR & 0.8505&0.8609  & 0.9518& 0.8637 &0.8187 &0.8691 \\ 
    DetectGPT &0.8517  & 0.8390 &0.9797 &0.8575  &0.8400 &0.8736  \\ 
    Fast-DetectGPT & 0.9834 &  0.9572&0.9984 &0.9592  &0.9404 &0.9677  \\ 
    TOCSIN & 0.9307& 0.9518 & 0.9188& 0.9424 &0.9357 &0.9359 \\ 
    \bottomrule
  \end{tabular*}
  \caption{AUROC of Likelihood, LogRank, LRR, DetectGPT, Fast-DetectGPT, and TOCSIN used as a standalone metric. The black-box setting is used for all zero-shot classifiers, with GPT-Neo-2.7 as surrogate model. The results are averaged across XSum, SQuAD, and WritingPrompts, with other settings identical to those in Table~\ref{tab:open-source-models}.
 }
  \label{tab:standalone}
\end{table*}

\begin{table*}[t]
  \centering\small
  \begin{tabular*}{1 \textwidth}{@{\extracolsep{\fill}}lcccccc}
    \toprule
    {\bf Method} & {\bf GPT-2} & {\bf OPT-2.7} & {\bf Neo-2.7} & {\bf GPT-J} & {\bf NeoX} & {\bf Avg.} \\
    \midrule
    \rowcolor{gray!50}
    \multicolumn{7}{c}{\textbf{The White-Box Setting}}\\
    LRR &0.9601  & 0.9401 &0.9522  &0.9179  & 0.8793 &0.9299  \\ 
    LRR+TOCSIN (GPTScore) & {\bf 0.9631} & {\bf 0.9517} & {\bf 0.9749} & {\bf 0.9342} & {\bf 0.9194} & {\bf 0.9487} \\
    {\it(Absolute $\uparrow$)} & {\it 0.30\%} & {\it 1.16\%} & {\it 2.27\%} & {\it 1.63\%} & {\it 4.01\%} & {\it 1.88\%} \\
    \hdashline
    Fast-DetectGPT & 0.9967 & 0.9908 & 0.9940 & 0.9866 & 0.9754 & 0.9887 \\ 
    Fast-DetectGPT+TOCSIN (GPTScore) & {\bf 0.9972} & {\bf 0.9918} & {\bf 0.9951} & {\bf 0.9880} & {\bf 0.9772} & {\bf 0.9899} \\
    {\it(Absolute $\uparrow$)} & {\it 0.05\%} & {\it 0.10\%} & {\it 0.11\%} & {\it 0.14\%} & {\it 0.18\%} & {\it 0.12\%} \\
    \rowcolor{gray!50}
    \multicolumn{7}{c}{\textbf{The Black-Box Setting}}\\
    LRR &0.8505  & 0.8609 &0.9518  &0.8637  &0.8187  &0.8691  \\ 
    LRR+TOCSIN (GPTScore) & {\bf 0.9242} & {\bf 0.9399} & {\bf 0.9749} & {\bf 0.9319} & {\bf 0.9097} & {\bf 0.9361} \\
    {\it(Absolute $\uparrow$)} & {\it 7.37\%} & {\it 7.90\%} & {\it 2.31\%} & {\it 6.82\%} & {\it 9.10\%} & {\it 6.70\%} \\
    \hdashline
    Fast-DetectGPT & 0.9834 & 0.9572 & 0.9984 & 0.9592 & 0.9404 & 0.9677 \\
    Fast-DetectGPT+TOCSIN (GPTScore) & {\bf 0.9859} & {\bf 0.9621} & {\bf 0.9988} & {\bf 0.9639} & {\bf 0.9476} & {\bf 0.9717} \\
    {\it(Absolute $\uparrow$)} & {\it 0.25\%} & {\it 0.49\%} & {\it 0.04\%} & {\it 0.47\%} & {\it 0.72\%} & {\it 0.40\%} \\
    \bottomrule
  \end{tabular*}
  \caption{AUROC of LRR, Fast-DetectGPT, and their +TOCSIN versions with token cohesiveness scores computed using GPTScore. During token cohesiveness calculation, the number of copies is fixed at $n=10$ and the token deletion proportion decreases to $\rho=0.1\%$. The results are averaged across XSum, SQuAD, and WritingPrompts, with other settings identical to those in Table~\ref{tab:open-source-models}.
 }
  \label{tab:GPTscore}
\end{table*}

\subsection{Details of Time \& Space Efficiency}\label{subsec:time-space}
Table~\ref{tab:time-space_details} reports detailed time/space analysis results of Fast-DetectGPT and Fast-DetectGPT+TOCSIN in black-box setting on XSum, SQuAD, and WritingPrompts, with passages generated by the five open-source LLMs. As TOCSIN always performs 10 rounds of random token deletion and uses BART-base to calculate token cohesiveness without any other requirements, the additional time/space costs are rather stable across datasets and source models (as long as the passages are roughly of equal size). On average, TOCSIN bring an additional runtime of 0.16s per instance and additional GPU memory usage of 4.71GB.

\subsection{TOCSIN as A Standalone Metric}\label{subsec:standalone}
We evaluate TOCSIN as a standalone metric and compare it with our baselines. Since TOCSIN is a fully black-box detector (which even does not require a surrogate model), we compare their performance under the black-box setting. Table~\ref{tab:standalone} shows the average AUROC across XSum, SQuAD, and WritingPrompts. The results reveal that TOCSIN, when used alone, outperforms all competitive baselines except the current best Fast-DetectGPT.

We further examine the score distributions to understand why it trails behind Fast-DetectGPT. We find that unlike Fast-DetectGPT, where humanwritten and LLM-generated text scores are almost separate with minimal overlap \cite[Figure 1]{bao2024fastdetectgpt}, TOCSIN scores for human-written text fall largely within the range of LLM-generated text scores , as shown in Figure~\ref{fig:example1}. This makes it particularly challenging to identify LLM-generated text with very low TOCSIN scores, as these scores fall perfectly within the range for human-written text.

Moreover, we would like to emphasize that TOCSIN’s lesser performance when used alone, compared to the current best Fast-DetectGPT, does not diminish its value as a plug-and-play module to enhance zero-shot detectors. As we have shown in Section~\ref{subsec:analyses}, TOCSIN’s unique strength lies in its ability to complement existing detectors, thereby providing improvements when combined.

\subsection{Impact of Semantic Difference Metric}\label{subsec:gptscore}
\begin{figure}[t]
  \centering
  \includegraphics[width=0.48\textwidth]{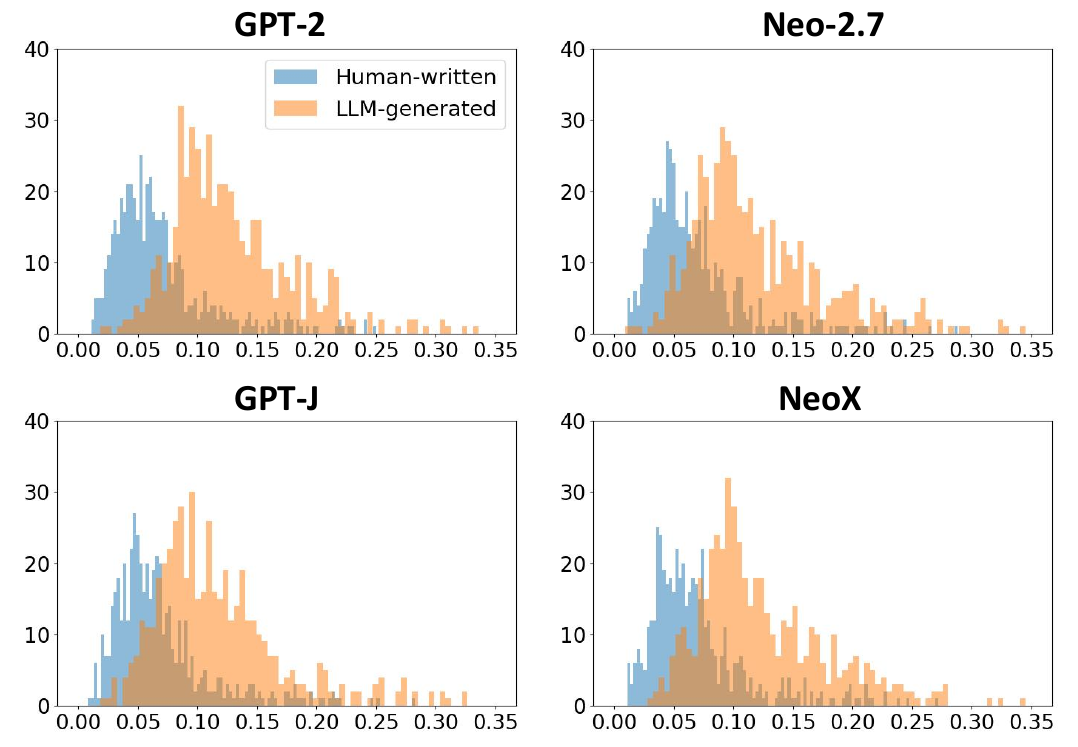}
  \caption{Distributions of token cohesiveness scores computed with GPTScore between human-written and LLM-generated articles. All the articles are identical to those used in Figure~\ref{fig:example1}.}
  \label{fig:cohesiveness-gptscore}
\end{figure}

TOCSIN requires a semantic difference metric for token cohesiveness calculation. Besides the negative BARTScore used in the main experiments, we further evaluate the negative GPTScore as another metric. We compute token cohesiveness scores using the new metric, and compare the distributions of the scores between LLM-generated and human-written passages. Figure~\ref{fig:cohesiveness-gptscore} visualizes the results for the same passages that were used in Figure~\ref{fig:example1}, showing that token cohesiveness scores computed using the new metric still exhibit clear distributional differences between the two types of text.

We further integrate these new token cohesiveness scores into LRR and Fast-DetectGPT, and evaluate their performance on XSum, SQuAD, and WritingPrompts. The results are given in Table~\ref{tab:GPTscore}, showing that, with the new metric GPTScore, TOCSIN can still bring consistent improvements to the base detectors. The absolute improvements in average AUROC reach 1.88\%/0.12\% in the white-box setting and 6.70\%/0.40\% in the black-box setting over LRR/Fast-DetectGPT, respectively.

\end{document}